%% file: main.tex
\documentclass[10pt,twocolumn,letterpaper]{article}

\usepackage[pagenumbers]{cvpr}

\input{preamble}

\definecolor{cvprblue}{rgb}{0.21,0.49,0.74}
\usepackage[pagebackref,breaklinks,colorlinks,allcolors=cvprblue]{hyperref}

\def\confName{CVPR}
\def\confYear{2027}

\title{One View, One Action: Spatially Generalizable VLAs through Observation-Aligned Experts and Cross Graph Consistency}

\author{
  Jinxi Liu$^{1}$\footnotemark[1] \quad
  Ruifeng Zhai$^{1}$\footnotemark[1] \quad
  Hao Liu$^{1}$ \quad
  Guangrun Wang$^{1,2}$\footnotemark[2] \\
  $^1$Sun Yat-sen University \quad
  $^2$X-Era AI Lab\\
  {\tt\small
    \{liujx233,liuh369\}@mail2.sysu.edu.cn,
    zhairf@mail.sysu.edu.cn, wanggrun@gmail.com
    }
}

\begin{document}
\maketitle

\renewcommand{\thefootnote}{\fnsymbol{footnote}} 
\footnotetext[1]{Equal contribution} 
\footnotetext[2]{Corresponding author}

\input{sec/0_abstract}    
\input{sec/1_introduction}

\input{sec/2_related}

\input{sec/3_method}

\input{sec/4_experiments}

\input{sec/5_conclusion}

{
    \small
    \bibliographystyle{ieeenat_fullname}
    \bibliography{main}
}

\newpage
\input{sec/x_appendix}

\end{document}

%% file: preamble.tex
\usepackage{url}
\usepackage{booktabs}
\usepackage{makecell}
\usepackage{tabularx}
\usepackage{array}
\usepackage{graphicx}
\usepackage{wrapfig}
\usepackage{multirow} 
\definecolor{gain}{RGB}{0,120,70}

%% file: sec/0_abstract.tex
\begin{abstract}
Multi-view vision-language-action (VLA) models still face challenges in spatial generalization, particularly under changes in camera viewpoints and robot initial states. These models typically receive action supervision only in the robot base frame, requiring them to implicitly learn the spatial correspondence between observations from each view and base-frame actions. We propose One View, One Action (OVOA-VLA), which learns to predict actions in each view's camera frame, allowing the same physical motion to provide diverse, observation-aligned supervision across views. OVOA-VLA jointly trains a base-frame expert and view-specific action experts, whose robot-state inputs and action targets are expressed in their respective camera frames.
Two complementary directed attention graphs model both directions of interaction between the base and view-specific experts, jointly leveraging view-specific action supervision to improve base-frame prediction. Cross Graph Consistency further encourages agreement between the two graphs' base-frame predictions for the same sample. With the same multi-view inputs, the model supports both full and base-only inference.
Experiments across multiple simulation benchmarks demonstrate improved manipulation performance and robustness to camera-viewpoint and robot-initialization perturbations. Full and base-only inference achieve average success rates of 88.1\% and 85.7\% on LIBERO-Plus, respectively. In real-world evaluations, base-only inference raises the average success rates from $\pi_{0.5}$'s 28.3\% to 58.3\% across three tasks under spatial distribution shifts.
\end{abstract}

%% file: sec/1_introduction.tex
\section{Introduction}
Vision-Language-Action (VLA) models transfer visual and semantic knowledge from large-scale pretrained vision-language models to robotic control, achieving substantial progress in language-conditioned manipulation~\citep{brohan2023rt, GR00T2025Johan, intelligence2025pi05visionlanguageactionmodelopenworld, kim2024openvla, pi0-BlackK-RSS-25, pertsch2025pi0fast,shukor2025smolvla}. 
However, reliable manipulation also requires spatial generalization: the ability to generate appropriate actions across changes in camera viewpoints and robot initial states. Viewpoint shifts alter the visual appearance of scenes and robot motion~\citep{li2024simpler}, whereas variations in the robot’s state may require different motion trajectories to accomplish the same task. Yet existing benchmarks reveal substantial performance degradation under both types of shifts, highlighting the persistent challenge of spatial generalization~\citep{fei25libero-plus,zhang2026world,RADAR2026chen, colosseumv22026Jeremy, colosseum2024Wilbert}.

\begin{figure*}[t]
    \centering
    \includegraphics[width=0.9\linewidth]{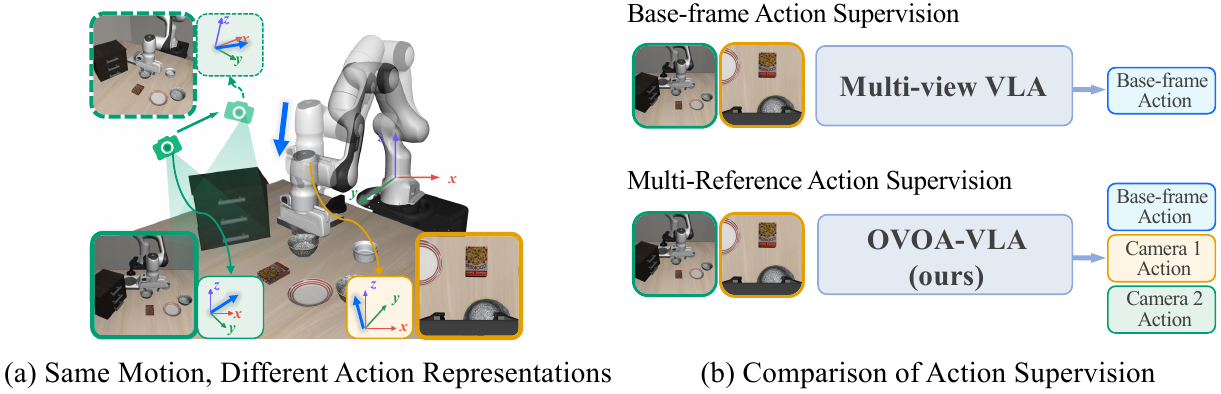}
    \vspace{-10pt}
    \caption{\small{\textbf{Motivation and comparison of action supervision.}
    (a) The same physical action has different representations in the robot base frame and different camera frames; as the camera pose changes, its representation in the corresponding camera frame changes accordingly.
    (b) Existing multi-view VLAs typically supervise actions only in the base frame. OVOA-VLA retains this supervision and adds action supervision in each view's camera frame using the same demonstration.}}
    \label{fig:teaser}
    \vspace{-13pt}
\end{figure*}

Action representation provides a useful perspective on this spatial generalization challenge. The same physical robot motion can have different numerical representations across coordinate frames~\citep{zhang2025-ocvla,zhang2026ocvlaplus, li2026fixedfreecamerascalibrationfree}. Figure~\ref{fig:teaser}(a) illustrates this for a single end-effector movement: its displacement components differ when expressed in the robot base coordinates and in each camera's coordinates. Yet multi-view VLAs typically combine observations from multiple cameras while receiving action supervision only in the robot base frame~\citep{kim2025finetuning}, requiring them to implicitly learn the spatial correspondence between observations from each view and base-frame actions~\citep{zhang2025-ocvla, jang2026axisguide}. Consequently, the learned correspondence may become tied to spatial configurations encountered during training, limiting generalization to unseen camera viewpoints and robot initial states~\citep{pmlr-v305-xing25a, tang2026movingeyeenhancingvla, gerigk2026listeningeasierscrubbingvisual}. 
Prior work has explored camera-frame action prediction~\citep{zhang2025-ocvla,li2026fixedfreecamerascalibrationfree} and explicit camera-geometry conditioning~\citep{jiang2026viewinvariant} to improve this spatial correspondence. These studies highlight the importance of reference-frame choice and camera geometry in action learning. This motivates jointly leveraging action supervision from all available views to improve the spatial generalization of the base-frame policy. As shown in Figure~\ref{fig:teaser}(b), we introduce camera-frame action supervision for each view while retaining the original base-frame supervision.

We introduce \textbf{One View, One Action (OVOA-VLA)}, a framework for learning \emph{observation-aligned actions}. Its organizing principle is to \textbf{predict actions expressed in the corresponding camera frame for each view}.  Specifically, we transform demonstration actions and robot states into each camera frame using known coordinate transformations. For each view, the model predicts actions in the corresponding frame, conditioned on the associated image and robot state expressed in that frame. In this way, the same physical motion provides diverse, observation-aligned action supervision across views, allowing each view to contribute to learning the base-frame policy not only through visual observations but also through its corresponding action prediction task. 
This encourages the policy to rely less on observation--action correspondences specific to the training configurations.

We instantiate OVOA-VLA with a Mixture of Transformers~\citep{liang2024mixture}, comprising a view-specific action expert for each input view and a base-frame expert. Each view-specific expert directly attends to visual tokens from its corresponding view, receives robot state expressed in that camera frame, and predicts actions in the same frame. The base expert uses all visual inputs and the base-frame robot state, while retaining the original base-frame action supervision.
We jointly train these experts under two complementary directed attention graphs. In the base-to-reference graph, the view-specific experts attend to action features from the base expert; in the reference-to-base graph, the base expert attends to action features from the view-specific experts. During training, we perform a forward pass under each graph for the same sample, with both passes sharing model parameters and receiving demonstrated-action supervision. Since both base-frame predictions correspond to the same action target, we introduce Cross Graph Consistency (CGC) to encourage agreement between them. Our model supports two inference modes: full inference uses the reference-to-base graph, while base-only inference uses the base-to-reference graph, where the base expert does not attend to the view-specific experts and thus permits their removal.

OVOA-VLA demonstrates strong performance across multiple simulation benchmarks~\citep{liu2023libero,fei25libero-plus,mandlekar2023mimicgen,Zhang_2025_ICCV}. Full inference achieves average success rates of 98.7\% and 88.1\% on LIBERO and LIBERO-Plus. On LIBERO-Plus, it improves success rates from 66.2\% to 75.4\% under camera viewpoint perturbations and from 67.2\% to 88.4\% under robot initialization perturbations, compared with $\pi_{0.5}$. Base-only inference achieves 85.7\% on LIBERO-Plus, demonstrating that the gains persist without view-specific experts. 
In real-world evaluations on a Franka Research 3 robot, base-only inference outperforms $\pi_{0.5}$ by 5.0 percentage points in average success rate across six in-distribution tasks and raises the average success rate from 28.3\% to 58.3\% across three tasks under spatial distribution shifts.
Our main contributions are as follows:
\begin{itemize}
    \item \textbf{Observation-aligned action learning.} We introduce OVOA-VLA, which jointly learns camera-frame actions from diverse, observation-aligned supervision derived from the same demonstration, improving base-frame policy generalization.

    \item \textbf{Complementary expert training.} We design two complementary directed attention graphs to model both interaction directions between the base and view-specific experts, with CGC encouraging consistent base-frame predictions across graphs for the same sample.

    \item \textbf{Two inference modes.} Our model supports full inference with the reference-to-base graph and base-only inference with the base-to-reference graph, where the base expert is independent of the view-specific experts.

    \item \textbf{Experimental validation.} Experiments across four simulation benchmarks and six real-world tasks demonstrate improved manipulation performance and robustness to camera-viewpoint and robot-initialization shifts.
\end{itemize}

%% file: sec/2_related.tex
\section{Related Work}
\begin{figure*}[t]
    \centering
    \includegraphics[width=\linewidth]{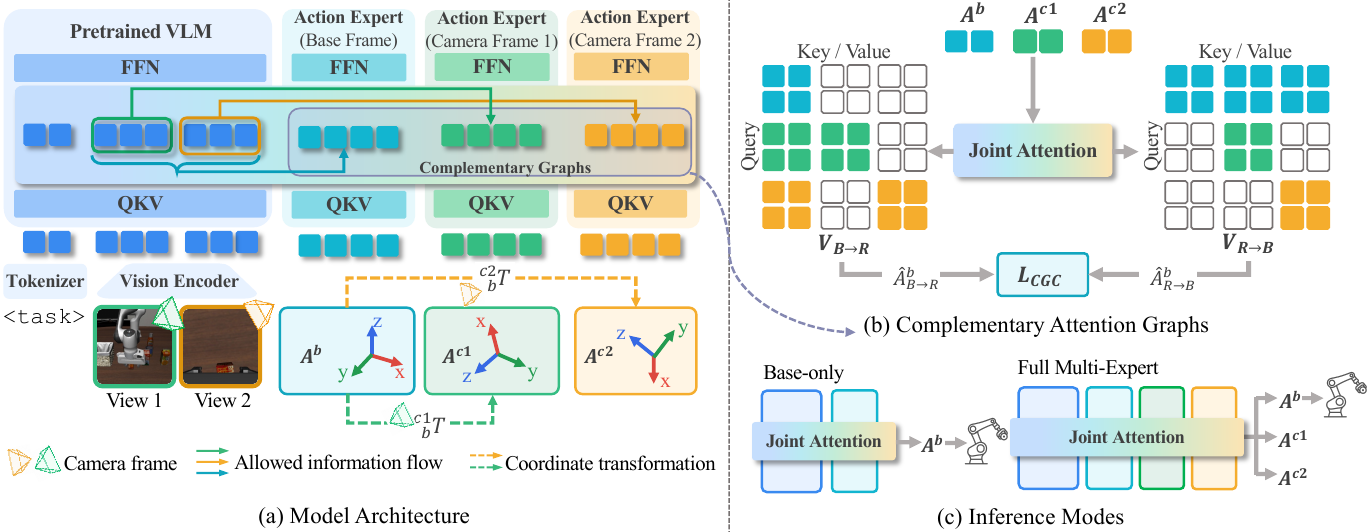}
    \vspace{-13pt}
    \caption{\textbf{Overview of OVOA-VLA.}(a) Each view-specific action expert attends to its own view and predicts camera-frame actions, jointly trained with the base expert. (b) Complementary attention graphs enable directed interactions: view-specific experts attend to base action features ($B\rightarrow R$), while the base expert attends to view-specific action features ($R\rightarrow B$). CGC encourages consistency between the two base-frame predictions. (c) Full Inference uses $R\rightarrow B$; Base-only Inference uses $B\rightarrow R$, omitting view-specific experts while retaining all input views. Both execute base-frame actions.
    }
    \vspace{-13pt}
    \label{fig:overview}
\end{figure*}

\paragraph{Vision-Language-Action Models}
Recent vision-language-action (VLA) models adapt pretrained vision-language models to robotic control by training on large-scale robot demonstrations for direct action prediction~\citep{pertsch2025pi0fast,kim2024openvla,brohan2023rt,pi0-BlackK-RSS-25}. These models have demonstrated strong performance in language-conditioned manipulation and generalization across objects, scenes, and robot embodiments~\citep{o2024open,kim2024openvla,pertsch2025pi0fast}. A major design choice in VLAs lies in how actions are encoded and generated, ranging from autoregressive prediction of discretized action tokens~\citep{kim2024openvla,brohan2023rt,pertsch2025pi0fast} to continuous action generation with diffusion~\citep{li2024cogact} or flow-matching objectives~\citep{pi0-BlackK-RSS-25}. Another important consideration is the reference frame in which actions are expressed. Many VLAs predict joint-space actions or Cartesian actions expressed in robot-base or world frames rather than the input camera frames~\citep{zhang2025-ocvla}, leaving the geometric correspondence between camera-centered visual observations and action targets to be learned implicitly.

\paragraph{Spatial Generalization in Vision-Language-Action Models}
Existing approaches to spatial generalization can be broadly grouped by their focus on observations, spatial perception, or action prediction. VISTA augments demonstrations with synthesized camera views, while Cai et al. study multi-view demonstrations and generate additional views using RoboNVS~\citep{tian2024view,cai2026viewpointgeneralizationmultiviewdemonstrations}. AnyCamVLA instead adapts test-time observations toward the training camera configuration without policy retraining~\citep{heo2026anycamvla}. For spatial perception, Spatial Forcing aligns visual features with pretrained 3D representations, QDepth-VLA introduces quantized depth prediction as auxiliary supervision, and ActiveVLA actively selects viewpoints and zooms into critical regions~\citep{li2025spatialforcingimplicitspatial,li2026qdepthvlaquantizeddepthprediction,liu2026activevla}. For action prediction, OC-VLA predicts camera-frame actions using camera extrinsics, whereas CamVLA jointly predicts camera-frame actions and hand-eye transformations for calibration-free control~\citep{zhang2025-ocvla,li2026fixedfreecamerascalibrationfree}. OC-VLA++ introduces geometry-guided cross-view action equivariance, while Huang et al. enforce consistency between flow velocity predictions from action-equivalent views~\citep{huang2026cross}. Qwen-RobotManip further adopts camera-frame EEF deltas with camera positional encoding~\citep{yuan2026qwen}. Our work is most closely related to action-level approaches, introducing observation-aligned action supervision as auxiliary prediction tasks for individual views while retaining the original base-frame action supervision.

%% file: sec/3_method.tex
\section{Method}

\subsection{Problem Formulation}
\label{sec:method}
We consider language-conditioned robotic manipulation with multi-view visual observations. At timestep $t$, a VLA policy $\pi_\theta$ parameterized by $\theta$ receives observations $\mathcal O_t=\{I_t^{c_i}\}_{i=1}^{M}$ from $M$ cameras, a language instruction $l$, and the current robot state $s_t^b=[p_t^b,r_t^b,q_t^g]$, where $p_t^b$ and $r_t^b$ denote the absolute end-effector position and orientation in the robot base frame $b$, and $q_t^g$ denotes the gripper state. The policy predicts an action chunk:
\begin{equation}
    A_t^b=
    \left[
    a_t^b,\ldots,a_{t+H-1}^b
    \right]
    =
    \pi_\theta(\mathcal{O}_t,l,s_t^b),
\end{equation}
where $H$ is the action horizon and $b$ denotes the executable robot control frame. 
Each action is represented as:
\begin{equation}
    a_t^b=
    \left[
    \Delta p_t^b,\Delta r_t^b,g_t
    \right],
\end{equation}
where $\Delta p_t^b$ and $\Delta r_t^b$ denote the translational displacement and relative rotation of the end effector, and $g_t$ is the gripper command. While robot demonstrations provide state and action supervision in this control frame, visual observations are captured from multiple camera reference frames.

\subsection{Observation-Aligned Action Learning}

\paragraph{Constructing Multi-Reference Action Supervision.}
To construct observation-aligned action supervision, we transform demonstrated robot states and actions into their corresponding camera reference frames using known coordinate transformations. For each input image $I_t^{c_i}\in\mathcal{O}_t$ at observation timestep $t$, let $c_i$ denote its camera frame, and ${}^{c_i}_{b}T_t$ denote the rigid transformation from the robot base frame $b$ to $c_i$, with ${}^{c_i}_{b}R_t$ as its rotational component. Given the base-frame robot state $s_t^b=[p_t^b,r_t^b,q_t^g]$, let $e$ denote the end-effector frame and ${}^{b}_{e}T_t$ its homogeneous transformation matrix constructed from the end-effector position $p_t^b$ and orientation $r_t^b$. The transformed pose is given by:
\begin{equation}
\begin{aligned}
{}^{c_i}_{b}T_t\,{}^{b}_{e}T_t
&=
{}^{c_i}_{b}T_t
\begin{bmatrix}
\mathcal R(r_t^b) & p_t^b\\
\mathbf{0}^{\top} & 1
\end{bmatrix}
=
\begin{bmatrix}
\mathcal R(r_t^{c_i}) & p_t^{c_i}\\
\mathbf{0}^{\top} & 1
\end{bmatrix}.
\end{aligned}
\end{equation}
Here, $\mathcal R(\cdot)$ converts an orientation representation into a rotation matrix, while $p_t^{c_i}$ and $r_t^{c_i}$ denote the end-effector position and orientation in the camera frame, respectively. Keeping the gripper state $q_t^g$ unchanged, we obtain the camera-frame robot state $s_t^{c_i}=[p_t^{c_i},r_t^{c_i},q_t^g]$.

For the base-frame action $a_t^b=[\Delta p_t^b,\Delta r_t^b,g_t]$, we further transform its translational and rotational components into the corresponding camera frame to construct action supervision targets. For a base-frame rotational increment encoded as a rotation vector, we first convert $\Delta r_t^b$ into the rotation matrix $\Delta R_t^b$. Using the rotational component ${}^{c_i}_{b}R_t$ of the previously defined transformation, we obtain:
\begin{equation}
\Delta p_t^{c_i}
=
{}^{c_i}_{b}R_t\,\Delta p_t^b,
\qquad
\Delta R_t^{c_i}
=
{}^{c_i}_{b}R_t\,\Delta R_t^b\,{}^{b}_{c_i}R_t.
\end{equation}
Here, ${}^{b}_{c_i}R_t=({}^{c_i}_{b}R_t)^\top$. We then convert $\Delta R_t^{c_i}$ back to the rotation-vector representation $\Delta r_t^{c_i}$, yielding $a_t^{c_i}=[\Delta p_t^{c_i},\Delta r_t^{c_i},g_t]$, while leaving the gripper command $g_t$ unchanged. Applying the corresponding coordinate transformations to all $H$ actions yields the camera-frame action chunk $A_t^{c_i}=[a_t^{c_i},\ldots,a_{t+H-1}^{c_i}]$. For wrist-mounted cameras, whose extrinsics vary over time, actions within a chunk are transformed using either the camera extrinsics at their corresponding timesteps or those at the observation timestep $t$ for the entire chunk, depending on the temporal reference convention of the action representation. Platform-specific action representations and temporal conventions are detailed in Appendix~\ref{app:state_action_representations}. Together with the transformed state $s_t^{c_i}$, this provides each view $I_t^{c_i}$ with frame-aligned state conditioning and action supervision. Thus, the same demonstrated motion provides diverse, observation-aligned action supervision across reference frames without additional action annotation, while retaining the original base-frame action supervision.

\paragraph{View-Specific Action Experts.}
To learn from the constructed multi-reference action supervision, we adopt a Mixture of Transformers (MoT) architecture~\cite{liang2024mixture} that retains the base-frame action expert and introduces an independently parameterized action expert for each camera view. All action experts share a vision-language backbone and are conditioned on the same language instruction $l$. The base-frame expert takes the robot state $s_t^b$ and predicts the action chunk $A_t^b$ used for robot execution, while each view-specific expert takes the corresponding camera-frame state $s_t^{c_i}$ and predicts $A_t^{c_i}$.

To align each view-specific expert's visual input with its action representation, we introduce \textbf{Reference-Aligned Visual Conditioning}. The expert associated with $c_i$ directly attends only to visual tokens from $I_t^{c_i}$, whereas the base-frame expert retains access to visual tokens from the complete multi-view observation $\mathcal{O}_t$.

\subsection{Action Expert Interaction and Consistency}

\paragraph{Complementary Attention Graphs.}
To enable the view-specific action prediction tasks to contribute to learning the base-frame policy, we introduce two complementary directed attention graphs that control cross-expert token visibility within the MoT self-attention layers. 
The two graphs share model parameters but use different attention masks to separately model the two directions of interaction between the base-frame and view-specific experts, rather than coupling them within a single bidirectional attention graph.
Here, $B$ denotes the base-frame action expert, while $R$ collectively denotes the view-specific experts associated with the camera reference frames, with arrows indicating the direction of information flow. In the \textbf{base-to-reference graph} ($B\rightarrow R$), the view-specific action tokens attend to the base-frame action tokens, while reverse attention is blocked. This allows the view-specific action losses to update the base expert through the attention connections, while keeping base-frame predictions independent of the view-specific action features. In the \textbf{reference-to-base graph} ($R\rightarrow B$), the attention direction is reversed, allowing the base-frame action tokens to attend to the view-specific action features and leverage them for base-frame action prediction.

Ordering the action experts as $(b,c_1,\ldots,c_M)$, we define the corresponding block visibility matrices as:
\begin{equation}
\mathbf{V}_{B\rightarrow R}
=
\begin{bmatrix}
1 & \mathbf{0}^{\top} \\
\mathbf{1} & \mathbf{I}_{M}
\end{bmatrix},
\qquad
\mathbf{V}_{R\rightarrow B}
=
\begin{bmatrix}
1 & \mathbf{1}^{\top} \\
\mathbf{0} & \mathbf{I}_{M}
\end{bmatrix}.
\end{equation}
Here, rows and columns correspond to query and key/value expert blocks, respectively, with entries of $1$ indicating permitted attention. The identity block $\mathbf{I}_{M}$ allows attention within each view-specific expert while blocking direct attention between different view-specific experts.

\paragraph{Cross Graph Consistency.}
The two attention graphs produce base-frame action predictions under different cross-expert dependencies, while both target the same demonstrated action. We therefore introduce Cross Graph Consistency (CGC) to encourage agreement between these predictions. During training, we perform two forward passes for each sample, one under each attention graph, while keeping the conditioning inputs, flow timestep, and per-expert noise samples identical across both passes. From the predicted flow velocities, we recover the clean base-frame action estimates, denoted as $\hat A^b_{B\rightarrow R}$ and $\hat A^b_{R\rightarrow B}$.

To establish a clear alignment direction for each consistency update, we formulate the loss with randomized stop-gradient, aligning one prediction with the detached prediction of the other branch rather than backpropagating through both branches. We sample $z\sim\operatorname{Bernoulli}(0.5)$ to select the detached branch with equal probability. The CGC loss is defined as:
\begin{equation}
\mathcal{L}_{\mathrm{CGC}}=
\begin{cases}
\operatorname{MSE}\left(
\hat A^b_{R\rightarrow B},
\operatorname{sg}[\hat A^b_{B\rightarrow R}]
\right), & z=0,\\[3pt]
\operatorname{MSE}\left(
\hat A^b_{B\rightarrow R},
\operatorname{sg}[\hat A^b_{R\rightarrow B}]
\right), & z=1.
\end{cases}
\end{equation}
Here, $\operatorname{sg}(\cdot)$ denotes the stop-gradient operator. Both branches retain ground-truth flow-matching supervision.

\subsection{Training and Inference}

\paragraph{Training Objective.}
We jointly train the action experts across reference frames using conditional flow matching. For an action chunk $A_t^f$ in frame $f\in\mathcal{F}=\{b,c_1,\ldots,c_M\}$, we independently sample Gaussian noise $\epsilon^f\sim\mathcal{N}(0,\mathbf{I})$ for each reference frame, while sharing the same flow timestep $\tau$ across all experts within a sample:
\begin{equation}
    A_{t,\tau}^f
    =
    (1-\tau)A_t^f+\tau\epsilon^f .
\end{equation}

Let $\mathcal{L}_{\mathrm{FM}}^f(\mathbf{V})$ denote the standard conditional flow-matching loss for the action expert in frame $f$ under the attention graph specified by $\mathbf{V}$. We jointly optimize the flow-matching losses of all action experts under both attention graphs together with the cross-graph consistency loss:
\begin{equation}
\begin{aligned}
\mathcal{L}_{\mathrm{train}}
={}&
\frac{1}{2}
\sum_{\mathbf{V}\in
\{\mathbf{V}_{B\rightarrow R},\,\mathbf{V}_{R\rightarrow B}\}}
\Bigl[
\mathcal{L}_{\mathrm{FM}}^b(\mathbf{V})\\
&\quad+
\lambda_{\mathrm{cam}}
\sum_{i=1}^{M}
\mathcal{L}_{\mathrm{FM}}^{c_i}(\mathbf{V})
\Bigr]
+
\lambda_{\mathrm{CGC}}\mathcal{L}_{\mathrm{CGC}}.
\end{aligned}
\end{equation}
The base-frame loss has unit weight, while each view-specific expert's loss is weighted by $\lambda_{\mathrm{cam}}=0.1$. We set the consistency weight to $\lambda_{\mathrm{CGC}}=0.02$.

\paragraph{Inference Modes.}
Our model supports two inference modes: \textbf{Full Multi-Expert Inference} and \textbf{Base-only Inference}, both using the base-frame action for robot execution. In Full Multi-Expert Inference, we adopt the reference-to-base graph ($R\rightarrow B$), allowing base-frame action tokens to attend to view-specific action tokens and thus requiring all action experts for prediction. In contrast, the base-to-reference graph ($B\rightarrow R$) keeps base-frame predictions independent of view-specific action-token features, enabling Base-only Inference with only the base-frame action expert while omitting the auxiliary view-specific experts. The selected inference mode remains fixed throughout the entire rollout. 
Base-only Inference therefore maintains inference cost comparable to the baseline with a single action expert, while Full Multi-Expert Inference leverages additional action features from the view-specific experts, providing a flexible trade-off between performance and computation.

%% file: sec/4_experiments.tex
\section{Simulation Experiments}
\subsection{Experimental Setup}

\paragraph{Simulation Benchmarks.}
We evaluate our method on four simulation benchmarks to assess both spatial generalization and broader manipulation capabilities: 
(i) \textbf{LIBERO}~\citep{liu2023libero} comprises four multi-task manipulation suites;
(ii) \textbf{LIBERO-Plus}~\citep{fei25libero-plus} extends LIBERO with controlled perturbations for evaluating out-of-distribution robustness, and we evaluate the LIBERO-trained model in a zero-shot manner;
(iii) \textbf{VLABench}~\citep{Zhang_2025_ICCV} is a large-scale benchmark for language-conditioned robotic manipulation, where we use Track 1 to evaluate downstream task learning and skill acquisition under in-distribution settings;
and (iv) \textbf{MimicGen}~\citep{mandlekar2023mimicgen} provides diverse manipulation tasks spanning long-horizon and high-precision behaviors, from which we select ten Panda D0 tasks to evaluate multi-task manipulation performance.

\paragraph{Baselines.}
On LIBERO~\citep{liu2023libero} and LIBERO-Plus~\citep{fei25libero-plus}, we compare against representative VLA baselines listed in Table~\ref{tab:libero_libero_plus}.
On VLABench~\citep{Zhang_2025_ICCV}, we use our reproduced $\pi_{0.5}$~\citep{intelligence2025pi05visionlanguageactionmodelopenworld} as the primary baseline and include results reported in the original benchmark paper for reference.
On MimicGen~\citep{mandlekar2023mimicgen}, we compare against our reproduced $\pi_{0.5}$~\citep{intelligence2025pi05visionlanguageactionmodelopenworld} under the same experimental protocol.

\paragraph{Implementation Details.}
We initialize our model from a pretrained $\pi_{0.5}$~\citep{intelligence2025pi05visionlanguageactionmodelopenworld} checkpoint and initialize all action experts with the weights of the original $\pi_{0.5}$ action expert.  
We train a separate OVOA-VLA model for each benchmark for 30,000 steps, using a batch size of 32 on two NVIDIA RTX PRO 6000D GPUs. 
At inference time, the policy predicts 10-step action chunks and replans after executing the first five actions. Additional preprocessing and augmentation details are provided in Appendix~\ref{app:details}.

\begin{table*}[t]
    \centering
    \caption{\small{\textbf{Performance comparison on LIBERO and LIBERO-Plus.} We report success rates (\%) across the four LIBERO task suites and their average. For LIBERO-Plus, we report Camera and Robot perturbation results and the overall success rate across all 10,030 evaluation instances. $^{*}$ denotes results reproduced by us. Complete results for all perturbation dimensions are provided in Appendix~\ref{app:libero_plus}.}}
    \label{tab:libero_libero_plus}
    \setlength{\tabcolsep}{3pt}
    \renewcommand{\arraystretch}{0.97}
    \vspace{-7pt}
    \resizebox{0.75\textwidth}{!}{
    \begin{tabular}{lccccc|ccc}
    \toprule
    \multirow{2}{*}{\textbf{Model}}
    & \multicolumn{5}{c|}{\textbf{LIBERO}}
    & \multicolumn{3}{c}{\textbf{LIBERO-Plus}} \\
    \cmidrule(lr){2-6}
    \cmidrule(lr){7-9}
    & \textbf{Spatial}
    & \textbf{Object}
    & \textbf{Goal}
    & \textbf{Long}
    & \textbf{Avg.}
    & \textbf{Camera}
    & \textbf{Robot}
    & \textbf{Total}
    \\
    \midrule
    UniVLA~\citep{bu2025univla}      & 96.5 & 96.8 & 95.6 & 92.0 & 95.2 & 1.8 & 46.2 & 42.9 \\
    WorldVLA~\citep{cen2025worldvla}    & 85.6 & 89.0 & 82.6 & 59.0 & 79.1 & 0.1 & 27.9 & 25.0 \\
    $\pi_{0}$-FAST~\citep{pertsch2025pi0fast}    & 96.4 & 96.8 & 88.6 & 60.2 & 85.5 & 65.1 & 21.6 & 61.6 \\
    OpenVLA-OFT~\citep{kim2025finetuning} & 97.6 & 98.4 & 97.9 & 94.5 & 97.1 & 56.4 & 31.9 & 69.6 \\
    DreamVLA~\citep{NEURIPS2025_22d4f952} & 97.5 & 94.0 & 89.5 & 89.5 & 92.6 & 65.0 & 40.9 & 69.9 \\
    MemoryVLA~\citep{ICLR2026_1fa3d6cc} & 98.4 & 98.4 & 96.4 & 93.4 & 96.7 & 42.7 & 44.9 & 70.2 \\
    Libra-VLA~\citep{wei-etal-2026-libra} & 98.6 & 99.4 & 98.0 & 92.8 & 97.2 & 68.9 & 48.8 & 79.5 \\
    VLA-JEPA~\citep{sun2026vla} & 96.2 & 99.6 & 97.2 & 95.8 & 97.2 & 63.3 & 67.1 & 78.0 \\   
    $\pi_{0}^{*}$~\citep{pi0-BlackK-RSS-25}   & 97.6 & 96.2 & 90.4 & 83.8 & 92.0 & 17.2 & 17.9 & 52.8 \\  $\pi_{0.5}^{*}$~\citep{intelligence2025pi05visionlanguageactionmodelopenworld} & 96.0 & 94.8 & 91.8 & 87.6 & 92.6 & 66.2 & 67.2 & 77.4 \\
    \midrule
    \textbf{Ours} (Base-only)& 96.4 & 98.6 & 95.6 & 94.4 & 96.3 & \textbf{78.4} & 81.2 & 85.7 \\
    \textbf{Ours} (Full) & \textbf{99.4} & \textbf{99.8} & \textbf{98.2} & \textbf{97.2} & \textbf{98.7} & 75.4 & \textbf{88.4} & \textbf{88.1} \\
    \bottomrule
\end{tabular}
}
\vspace{-10pt}
\end{table*}

\subsection{Results on LIBERO and LIBERO-Plus}
As shown in Table~\ref{tab:libero_libero_plus}, our model with Full Multi-Expert Inference achieves the best performance across all four LIBERO task suites, with an average success rate of 98.7\%, outperforming the strongest baseline by 1.5 percentage points. The improvement is particularly evident on the more challenging LIBERO-Long suite, where our method reaches 97.2\%.
On LIBERO-Plus, the full mode achieves an overall success rate of 88.1\%, exceeding the $\pi_{0.5}^{*}$ baseline by 10.7 percentage points. Furthermore, it outperforms this baseline by 9.2 and 21.2 percentage points under the Camera and Robot perturbations. Notably, under Base-only Inference, our model uses only the base-frame action expert and still achieves an overall success rate of 85.7\% on LIBERO-Plus, with 78.4\% and 81.2\% under Camera and Robot perturbations. Meanwhile, its inference cost remains comparable to that of the baseline with a single action expert.

\subsection{Evaluation across Diverse Manipulation Benchmarks}

\begin{table}[t]
    \centering
    \caption{\small{
        \textbf{Results on the VLABench Track-1 in-distribution benchmark.}
        We report the average success rate (SR), progress score (PS), and intention score (IS) over 10 tasks. $^{*}$ denotes results reproduced by us. Complete results are provided in Appendix~\ref{app:another_bench}.
    }}
    \vspace{-7pt}
    \label{tab:vlabench_track1_main}
    \small
    \setlength{\tabcolsep}{3pt}
    \renewcommand{\arraystretch}{0.95}
    \resizebox{\columnwidth}{!}{
        \begin{tabular}{lccc}
            \toprule
            \textbf{Method}
            & \textbf{Avg. SR $\uparrow$}
            & \textbf{Avg. PS $\uparrow$}
            & \textbf{Avg. IS $\uparrow$} \\
            \midrule
            $\pi_{0}$       & 47.0 & --   & --   \\
            $\pi_{0}$-FAST  & 51.2 & --   & --   \\
            $\pi_{0.5}^{*}$ & 42.2 & 57.1 & 70.4 \\
            \midrule
            \textbf{Ours} (Base-only) & 56.8 & 69.8 & 74.6 \\
            \textbf{Ours} (Full)      & \textbf{60.6} & \textbf{74.2} & \textbf{77.6} \\
            \bottomrule
        \end{tabular}
    }
    \vspace{-10pt}
\end{table}

\begin{table}[t]
    \centering
    \caption{\small{
        \textbf{Results on MimicGen Panda10 under the D0 setting.}
        We report the average success rate (SR) over 10 tasks with 100 episodes per task. Complete results are provided in Appendix~\ref{app:another_bench}.
    }}
    \vspace{-7pt}
    \label{tab:mimicgen_d0}
    \small
    \renewcommand{\arraystretch}{0.97}
    \begin{tabular*}{0.95\columnwidth}{@{\extracolsep{\fill}}lc@{}}
        \toprule
        \textbf{Method} & \textbf{Avg. SR (\%) $\uparrow$} \\
        \midrule
        $\pi_{0.5}^{*}$ & 64.4 \\
        \midrule
        \textbf{Ours} (Base-only) & 72.7 {\textcolor{gain}{(+8.3)}} \\
        \textbf{Ours} (Full) & \textbf{75.3} {\textcolor{gain}{(+10.9)}} \\
        \bottomrule
    \end{tabular*}
    \vspace{-10pt}
\end{table}

We further evaluate our method on \textbf{VLABench}~\citep{Zhang_2025_ICCV} and \textbf{MimicGen}~\citep{mandlekar2023mimicgen}. Compared with the common two-view input setting, VLABench provides three camera views, offering a natural setting to further examine the scalability of multi-reference action modeling with respect to the number of viewpoints. As shown in Table~\ref{tab:vlabench_track1_main}, both inference modes outperform the baseline, demonstrating improvements in task completion, execution progress, and target selection across diverse manipulation tasks.On MimicGen, both inference modes outperform the $\pi_{0.5}$ baseline on average, as shown in Table~\ref{tab:mimicgen_d0}.
Our method achieves clear improvements on long-horizon tasks such as Coffee Preparation and high-precision tasks such as Threading, indicating that multi-reference action modeling improves complex sequential execution and fine-grained manipulation.

\subsection{Ablations and Analysis}
\paragraph{Ablations on Method Design.}

\begin{table}[t]
    \centering
    \caption{\small{\textbf{Method design ablations on LIBERO and LIBERO-Plus.} We report success rates (\%); Camera and Robot denote the corresponding perturbations in LIBERO-Plus.}}
    \label{tab:ablation_libero}
    \vspace{-7pt}
    \small
    \setlength{\tabcolsep}{3pt}
    \renewcommand{\arraystretch}{1.05}
    \resizebox{\columnwidth}{!}{
        \begin{tabular}{lccc}
            \toprule
            Method & LIBERO Avg. & Camera & Robot \\
            \midrule
            Baseline ($\pi_{0.5}$) & 92.6 & 66.2 & 67.2 \\
            + View-Specific Action Experts & 96.5 & 73.1 & 71.6 \\
            + Ref.-Aligned Visual Cond. & 96.4 & 75.7 & 79.6 \\
            \midrule
            $B\!\rightarrow\!R$ Only & 96.6 & \textbf{79.1} & 77.0 \\
            $R\!\rightarrow\!B$ Only & 98.1 & 70.0 & 85.4 \\
            Bidirectional Attention & 95.2 & 77.1 & 71.7 \\
            Complementary Graphs & 98.3 & 74.1 & 86.8 \\
            + CGC (\textbf{Ours}) & \textbf{98.7} & 75.4 & \textbf{88.4} \\
            \bottomrule
        \end{tabular}
    }
    \vspace{-10pt}
\end{table}

We use $\pi_{0.5}$ as the baseline and first add view-specific action experts while retaining the base expert. At this stage, all experts share the same visual and language conditioning and independently predict actions expressed in their respective coordinate frames, without attending to one another. As shown in Table~\ref{tab:ablation_libero}, adding these experts improves the success rates on LIBERO, Camera, and Robot to 96.5\%, 73.1\%, and 71.6\%, respectively, indicating that supervision from the view-specific prediction tasks benefits both in-distribution performance and robustness to spatial perturbations. Building on this, Reference-Aligned Visual Conditioning further improves Camera and Robot success rates by 2.6 and 8.0 percentage points, respectively, while maintaining comparable performance on LIBERO. This suggests that aligning visual conditioning with the action reference frame helps improve action prediction under spatial distribution shifts.

We next compare attention graphs between the base expert and the view-specific experts. Each directed graph reveals complementary strengths: $B\!\rightarrow\!R$ Only reaches 79.1\% under Camera perturbations, whereas $R\!\rightarrow\!B$ Only achieves 85.4\% under Robot perturbations.
This complementarity also helps explain the inference-mode trade-off. Under Camera perturbations, Base-only Inference retains the benefits of multi-reference training without directly depending on view-specific action features, potentially reducing sensitivity to viewpoint-dependent variations. In contrast, Robot perturbations change the motion required from the robot. Full Multi-Expert Inference can leverage additional action features from different reference frames to better adapt to these changes.
We also evaluate Bidirectional Attention, which allows both directions within a single attention graph. However, it fails to preserve the respective advantages of the two unidirectional graphs, suggesting that simply coupling both directions is insufficient to fully exploit their complementarity. Based on this observation, we jointly train the two directed graphs (Complementary Graphs) to leverage both interaction directions, reaching 74.1\% and 86.8\% under Camera and Robot perturbations. Further adding Cross Graph Consistency (CGC) brings consistent gains across all evaluation settings, improving Camera and Robot success rates by 1.3 and 1.6 percentage points, respectively, while reaching 98.7\% on LIBERO. These results demonstrate the effectiveness of CGC in jointly training the two attention graphs.

\paragraph{Inference Efficiency.}
\begin{table}[t]
    \centering
    \caption{\small{\textbf{Performance and inference efficiency on LIBERO.} We compare the average success rate (\%), computational cost per action chunk (GFLOPs), peak GPU memory usage (GB), and mean inference latency (ms) across inference modes. Lower FLOPs, memory usage, and latency indicate better efficiency.}}
    \label{tab:inference_efficiency}
    \vspace{-7pt}
    \small
    \setlength{\tabcolsep}{3pt}
    \renewcommand{\arraystretch}{1}
    \resizebox{\columnwidth}{!}{
        \begin{tabular}{lcccc}
            \toprule
            Mode
            & Avg.\ $\uparrow$
            & FLOPs $\downarrow$
            & Mem.\ $\downarrow$
            & Lat.\ $\downarrow$ \\
            \midrule
            Baseline ($\pi_{0.5}$)
            & 92.6 & \textbf{254.65} & 10.56 & 63.07 \\
            \midrule
            $R \rightarrow B$
            & \textbf{98.7} & 263.79 & 19.12 & 103.07 \\
            $B \rightarrow R$
            & 97.0 & 263.79 & 19.10 & 99.97 \\
            Base-only
            & 96.3 & 254.66 & \textbf{10.49} & \textbf{62.77} \\
            \bottomrule
        \end{tabular}
    }
    \vspace{-13pt}
\end{table}
As shown in Table~\ref{tab:inference_efficiency}, both OVOA-VLA inference modes outperform the $\pi_{0.5}$ baseline on LIBERO. Full Multi-Expert Inference runs the view-specific action experts in addition to the base expert, increasing peak memory usage and mean inference latency by approximately 81\% and 63\%. Nevertheless, the FLOPs remain similar across inference modes because computation is dominated by the shared visual encoder and PaliGemma prefix encoding, while the action experts process only short token sequences.
Under $B\!\rightarrow\!R$, the base expert does not attend to the view-specific experts. Base-only Inference therefore omits those experts, achieving an increase of \textbf{3.7} percentage points while keeping FLOPs, peak memory usage, and inference latency comparable to $\pi_{0.5}$. All efficiency metrics are measured on the same NVIDIA RTX PRO 6000D GPU with a batch size of 1 and 10 denoising steps, with latency averaged over 100 inference runs.

\section{Real-World Experiments}
\begin{figure*}[htbp]
    \centering
    \includegraphics[width=\textwidth]
    {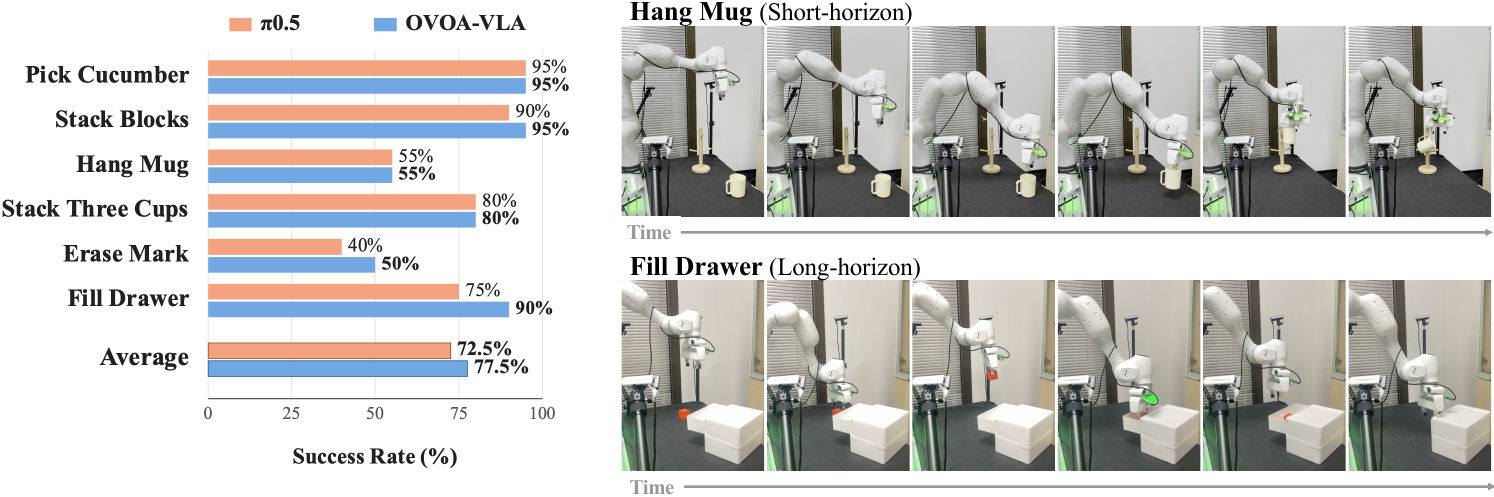}
    \vspace{-13pt}
    \caption{\small{
        \textbf{Real-world manipulation evaluation.} Left: in-distribution success rates across six tasks, with 20 rollouts per task. Right: representative successful rollouts of the short-horizon Hang Mug task and the long-horizon Fill Drawer task.}
    }
    \vspace{-5pt}
    \label{fig:real_world_results}
\end{figure*}

\begin{figure*}[htbp]
    \centering
    \includegraphics[width=\textwidth]
    {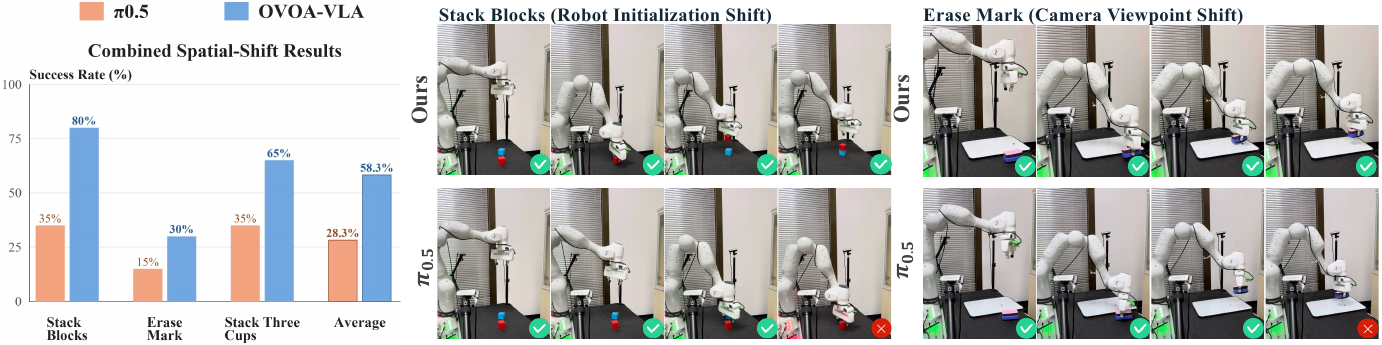}
    \vspace{-13pt}
    \caption{\small{\textbf{Zero-shot robustness under spatial shifts.}
    Left: task success rates pooled across the two shift types, with 20 rollouts per task.
    Middle and right: representative rollouts in which OVOA-VLA succeeds and $\pi_{0.5}$ fails. Shift-specific results are shown in Figure~\ref{fig:real_world_ood_shift_breakdown}.}
    }
    \vspace{-7pt}
    \label{fig:real_world_compare}
\end{figure*}
\subsection{Experimental Setup.}
We conduct real-world experiments on a 7-DoF Franka Research 3 (FR3) robot. 
The policy takes RGB images from a fixed third-person camera and a wrist-mounted camera, together with a task instruction, and predicts 6-DoF end-effector (EEF) target poses relative to the observation-time EEF pose, along with continuous gripper-closedness values.
The evaluation includes three short-horizon tasks and three long-horizon tasks, covering basic manipulation skills and multi-stage manipulation, respectively.
For each task, we collect 50 demonstration trajectories using GELLO teleoperation. During data collection, object initial positions are randomized within predefined ranges, while the third-person camera remains fixed. Demonstrations from all tasks are pooled to jointly post-train a single multi-task policy. For all real-world evaluations, we use Base-only Inference, which activates only the base-frame action expert, resulting in a single-expert inference setting comparable to that of $\pi_{0.5}$. Each method is evaluated over 20 rollouts per task, and we report task success rates. Detailed task definitions, hardware configurations, camera calibration, and training and execution settings are provided in Appendix~\ref{app:real_implementation}.

\subsection{Real-World Manipulation Evaluation}
Figure~\ref{fig:real_world_results} presents the in-distribution results and representative rollouts across six real-world manipulation tasks. OVOA-VLA achieves an average success rate of 77.5\%, surpassing $\pi_{0.5}$ at 72.5\% by 5.0 percentage points. It matches the baseline on three tasks and performs better on Stack Blocks, Erase Mark, and Fill Drawer, with gains of up to 15.0 percentage points. On the long-horizon tasks, OVOA-VLA improves the average success rate from 65.0\% to 73.3\%, with more pronounced gains on complex, multi-stage manipulation. The right panel shows representative successful rollouts of the short-horizon Hang Mug task and the long-horizon Fill Drawer task.

\subsection{Robustness under Spatial Distribution Shifts}
To further evaluate real-world spatial robustness, we select three tasks and consider two types of spatial distribution shifts. For Camera Viewpoint Shift, we evaluate two unseen fixed-camera views while retaining the same wrist-mounted camera. For Robot Initialization Shift, we define two perturbed initial EEF poses while keeping all other conditions unchanged. For each task and each configuration, we conduct five real-world rollouts. All OOD experiments are evaluated zero-shot without retraining or adaptation, and detailed task selection, perturbation settings, and shift-specific results are provided in Appendix~\ref{app:spatial_shift}.
Figure~\ref{fig:real_world_compare} shows that OVOA-VLA outperforms $\pi_{0.5}$ on all three tasks, increasing the overall success rate from 28.3\% to 58.3\%. This consistent improvement demonstrates stronger zero-shot robustness to spatial perturbations.

%% file: sec/5_conclusion.tex
\section{Conclusion}
We introduce OVOA-VLA, a framework that leverages observation-aligned action supervision across multiple reference frames to improve spatial generalization in VLA policies. By combining view-specific action experts with Complementary Attention Graphs and Cross Graph Consistency, our method enables action prediction tasks in different reference frames to jointly contribute to learning the executable base-frame policy. Experiments across four simulation benchmarks and six real-world manipulation tasks demonstrate improved manipulation performance and stronger robustness to camera-viewpoint and robot-initialization shifts. Notably, Base-only Inference retains performance gains without the view-specific action experts, while maintaining inference cost comparable to the baseline on LIBERO. These results highlight the value of leveraging multiple views not only as visual observations but also as sources of observation-aligned action supervision.

%% file: sec/x_appendix.tex
\appendix
\onecolumn
\section*{Supplementary Material}  
\section{Additional Related Work}
\paragraph{Mixture-of-Transformers for Multimodal Modeling}
Mixture-of-Transformers (MoT) employs modality-specific Transformer parameters while maintaining global self-attention over a joint token sequence, enabling specialized computation alongside cross-modal interaction~\citep{liang2024mixture}. This architecture has been applied across diverse multimodal tasks. BAGEL employs separate experts for multimodal understanding and generation~\citep{deng2025bagel}, while Mural connects a frozen LLM with an image diffusion model for text-to-image generation~\citep{jain2026mural}. Video-As-Prompt introduces a trainable MoT expert to guide a frozen video diffusion model for controllable video generation~\citep{bian2025videoasprompt}. In robotics, $\pi_0$~\citep{pi0-BlackK-RSS-25} and $\pi_{0.5}$~\citep{intelligence2025pi05visionlanguageactionmodelopenworld} couple a VLM with a separately parameterized action expert, while Motus integrates visual-understanding, video-generation, and action experts within a three-expert MoT architecture~\citep{bi2026motus}. In our work, we adopt MoT to accommodate multiple reference-frame action experts and their attention-based interactions.

\section{Action Representation Details}
\label{app:state_action_representations}

Table~\ref{tab:state_action_representations} summarizes the proprioceptive states and action representations used across platforms. LIBERO and MimicGen retain the native seven-dimensional incremental control commands, comprising three translational components, three rotational components in rotation-vector form, and one gripper command, with the original controller scales preserved. For VLABench and real-world Franka, absolute target poses are encoded relative to the end-effector pose observed at the start of the action chunk. VLABench uses position differences and component-wise XYZ Euler-angle differences, whereas Franka uses position differences and a rotation-vector representation of the relative orientation. All targets within a chunk share this initial pose as their reference. Gripper components retain their platform-specific definitions; for Franka, gripper closedness ranges from $0$ (fully open) to $1$ (fully closed). Actions are executed in the robot base frame on all platforms. For LIBERO and MimicGen, denormalized predictions are passed directly to the environment; for VLABench and Franka, predictions are first decoded into absolute target poses and then passed to the respective execution interfaces.

When constructing wrist-frame action supervision, we adopt temporal reference conventions consistent with each platform's action definition. For LIBERO and MimicGen, each action in a chunk is transformed using the wrist-camera extrinsics at its corresponding timestep. For VLABench and real-world Franka, the entire chunk uses the wrist-camera extrinsics at the initial observation timestep, which are updated upon replanning. 

\begin{table*}[htbp]
\centering
\caption{\textbf{Proprioceptive state and action representations.} Dimensions exclude padding. Numbers in parentheses indicate component dimensions. EEF denotes the end effector.}
\label{tab:state_action_representations}
\small
\setlength{\tabcolsep}{4pt}
\renewcommand{\arraystretch}{1.15}
\begin{tabularx}{\textwidth}{@{}l l c >{\raggedright\arraybackslash}X >{\raggedright\arraybackslash}X >{\raggedright\arraybackslash}X@{}}
\toprule
\textbf{Platform} & \textbf{Type} & \textbf{Dim.} & \textbf{Position} & \textbf{Orientation} & \textbf{Gripper} \\
\midrule
LIBERO / MimicGen & State & 8 & EEF position (3) & Rotation vector (3) & Finger joint positions (2) \\
& Action & 7 & Translational increment command (3) & Rotation-vector increment command (3) & Gripper command (1) \\
\addlinespace[5pt]
VLABench & State & 7 & EEF position (3) & XYZ Euler angles (3) & Gripper state (1) \\
& Action & 7 & Target position offset (3) & Component-wise XYZ Euler-angle difference (3) & Gripper command (1) \\
\addlinespace[5pt]
Real-world Franka & State & 8 & EEF position (3) & Quaternion in $wxyz$ order (4) & Gripper closedness (1) \\
& Action & 7 & Target position offset (3) & Relative rotation vector (3) & Target gripper closedness (1) \\
\bottomrule
\end{tabularx}
\end{table*}

\section{Additional Details of Simulation Experiments}
\label{app:simulation}
\subsection{Benchmark and Dataset Details}
\label{app:benchmarks}
\paragraph{LIBERO and LIBERO-Plus.}
The four task suites in LIBERO~\citep{liu2023libero} emphasize spatial relationships, manipulated objects, task goals, and long-horizon manipulation, respectively, providing a common environment for evaluating policies across different task requirements. We use all 40 tasks from the Spatial, Object, Goal, and Long suites and jointly train a single multi-task policy on the publicly released LIBERO demonstrations, with 50 trajectories per task and 2,000 trajectories in total. The original images have a resolution of $128\times128$, and we do not filter out failed trajectories before training. Each task is evaluated over 50 rollouts. LIBERO-Plus~\citep{fei25libero-plus} organizes perturbations into dimensions, enabling separate assessments of policy sensitivity to different environmental changes. We evaluate all seven official perturbation dimensions, following the official evaluation configurations and rollout counts. All LIBERO-Plus evaluations are conducted zero-shot using the same LIBERO-trained model, without training or adaptation on any perturbed data.

\paragraph{VLABench.}
VLABench~\citep{Zhang_2025_ICCV} is a benchmark for language-conditioned robotic manipulation, with multiple evaluation tracks assessing task learning and generalization. We adopt the in-distribution setting of Track 1, which evaluates the ability of policies to learn downstream tasks and acquire manipulation skills from demonstrations. We use the 10 primitive tasks in the fine-tuning setup, with 500 demonstration trajectories per task and 5,000 trajectories in total, and jointly train a single multi-task policy across all tasks. Visual observations comprise two fixed camera views and one wrist-mounted camera view, each with a resolution of $480\times480$. The robot state includes the end-effector position and orientation, along with the gripper state. Evaluation follows the official Track 1 task configurations, with 50 rollouts per task. We report success rate (SR), progress score (PS), and intention score (IS), averaged over the 10 tasks. SR measures overall task completion, PS measures execution progress as the proportion of completed subtasks, and IS reflects the tendency to select the correct target object based on whether the end effector approaches it. We treat SR as the primary evaluation metric, while PS and IS provide complementary measures for understanding task progress and target selection.

\paragraph{MimicGen.}
MimicGen~\citep{mandlekar2023mimicgen} is a framework for generating robotic manipulation demonstrations by transforming and adapting existing trajectories to different tasks and scenes. Its tasks include both long-horizon, multi-stage manipulation and precision-demanding assembly. For example, Coffee Preparation involves multiple stages, whereas Threading requires precise alignment and insertion. We select 10 D0 tasks using the Franka Emika Panda robot from the publicly released core dataset: Stack, Square, Coffee, Threading, Stack Three, Three Piece Assembly, Hammer Cleanup, Mug Cleanup, Coffee Preparation, and Kitchen. D0 denotes the default object initialization distribution, which we use for both training and evaluation. We jointly train a single multi-task policy using 1,000 demonstration trajectories per task, totaling 10,000 trajectories. Observations include third-person and wrist-camera images rendered at $224\times224$ resolution, together with the robot state.
We establish a unified evaluation protocol for these tasks, with 100 rollouts per task and the same pregenerated, fixed simulator initial states across all methods. We use the environments' native success conditions and terminate each rollout upon success or when the maximum episode length is reached. The maximum episode length is 400 steps for Stack, Square, Coffee, Threading, and Stack Three; 500 steps for Three Piece Assembly, Hammer Cleanup, and Mug Cleanup; and 800 steps for Coffee Preparation and Kitchen.

\subsection{Training and Implementation Details}
\label{app:details}
\paragraph{Training Configuration.}
We use the AdamW optimizer with $\beta_1=0.9$, $\beta_2=0.95$, a weight decay of $10^{-10}$, and a global gradient norm clipping threshold of 1.0. For LIBERO and MimicGen, the learning rate is linearly warmed up to $5\times10^{-5}$ over the first 10,000 updates and remains constant thereafter. For VLABench, the learning rate is linearly warmed up to $2.5\times10^{-5}$ over the first 1,000 updates, followed by cosine decay to $2.5\times10^{-6}$. All images are resized to $224\times224$ before being fed into the model. During training, we apply random color jitter to images from all camera views, varying brightness, contrast, and saturation to diversify visual appearance and reduce reliance on fixed lighting and color conditions. The corresponding augmentation strengths are set to 0.3, 0.4, and 0.5, respectively. For the third-person views in LIBERO and MimicGen and the primary fixed-camera view in VLABench, we additionally apply random cropping and in-plane rotation. The crop spans 95\% of the input width and height and is resized back to the original input dimensions, while the rotation angle is sampled from $[-5^\circ,5^\circ]$. These random augmentations are enabled only during training. We train the $\pi_{0.5}$ baseline on all three datasets using the same training data, optimizer, learning-rate schedule, and image preprocessing and augmentation settings as our method.

\paragraph{Software and Simulation Environments.}
We use JAX/Flax for model training and inference, with separate software environments for each simulation benchmark. LIBERO and LIBERO-Plus both use MuJoCo 3.2.3, paired with robosuite 1.4.1 and 1.4.0, respectively. VLABench uses MuJoCo 3.2.2. The MimicGen environment uses MimicGen 1.0.1, robosuite 1.4.1, robomimic 0.3.1, and MuJoCo 2.3.2.

\subsection{Complete Results on LIBERO-Plus}
Table~\ref{tab:libero_plus_full} presents the results across all seven perturbation dimensions. Beyond Camera and Robot, both inference modes outperform the reproduced $\pi_{0.5}$ baseline on Language, Light, Background, Noise, and Layout, showing that the improvements extend to a broader range of perturbations. Full mode demonstrates strong performance across all five dimensions, highlighting its broad robustness to variations in language, lighting, background, noise, and layout. Base-only mode also maintains strong robustness and further outperforms Full mode under camera-viewpoint perturbations.

\label{app:libero_plus}
\begin{table*}[t]
\centering
\caption{\textbf{Performance comparison on LIBERO-Plus.} We report success rates (\%) under all seven perturbation dimensions, together with the overall success rate. $^{*}$ denotes results reproduced by us.}
\label{tab:libero_plus_full}
\setlength{\tabcolsep}{3pt}
\renewcommand{\arraystretch}{1.05}
\begin{tabular}{lcccccccc}
    \toprule
    \textbf{Method} & \textbf{Camera} & \textbf{Robot} & \textbf{Language} & \textbf{Light} & \textbf{Background} & \textbf{Noise} & \textbf{Layout} & \textbf{Total} \\
    \midrule
    UniVLA & 1.8 & 46.2 & 69.6 & 69.0 & 81.0 & 21.2 & 31.9 & 42.9 \\
    WorldVLA & 0.1 & 27.9 & 41.6 & 43.7 & 17.1 & 10.9 & 38.0 & 25.0 \\
    $\pi_{0}$-FAST & 65.1 & 21.6 & 61.0 & 73.2 & 73.2 & 74.4 & 68.8 & 61.6 \\
    OpenVLA-OFT & 56.4 & 31.9 & 79.5 & 88.7 & 93.3 & 75.8 & 74.2 & 69.6 \\
    DreamVLA & 65.0 & 40.9 & 63.5 & 85.7 & 82.7 & 85.0 & 74.0 & 69.9 \\
    MemoryVLA & 42.7 & 44.9 & 84.4 & 92.8 & 95.0 & 62.1 & 84.7 & 70.2 \\
    Libra-VLA & 68.9 & 48.8 & 92.7 & \textbf{97.9} & 93.4 & 86.3 & 77.5 & 79.5 \\
    VLA-JEPA & 63.3 & 67.1 & \textbf{85.4} & 95.6 & 93.6 & 66.3 & 85.1 & 78.0 \\
    $\pi_{0}^{*}$ & 17.2 & 17.9 & 69.3 & 90.1 & 88.1 & 75.7 & 31.9 & 52.8 \\
    $\pi_{0.5}^{*}$ & 66.2 & 67.2 & 75.1 & 91.7 & 90.3 & 78.1 & 81.0 & 77.4 \\
    \midrule
    \textbf{Ours} (Base-only) & \textbf{78.4} & 81.2 & 82.6 & 94.6 & 94.0 & 88.8 & 85.6 & 85.7 \\
    \textbf{Ours} (Full) & 75.4 & \textbf{88.4} & 84.8 & 97.7 & \textbf{95.9} & \textbf{90.1} & \textbf{89.4} & \textbf{88.1} \\
    \bottomrule
\end{tabular}
\end{table*}

\subsection{Analysis of Full and Base-only Inference under Camera and Robot Perturbations}
\begin{table*}[t]
    \centering
    \caption{
        Success rates (\%) when executing camera-frame and wrist-frame
        actions under the $B \rightarrow R$ and $R \rightarrow B$
        attention graphs. Before execution, the predicted actions are
        transformed into the robot base frame using the corresponding
        camera extrinsics. Bold values indicate the better result between
        camera-frame and wrist-frame actions under the same attention graph
        for each perturbation type.
    }
    \label{tab:action_frame_attention}
    \setlength{\tabcolsep}{6pt}
    \begin{tabular}{@{}llccccccc@{}}
        \toprule
        \multirow{2}{*}{\textbf{Attention graph}}
        & \multirow{2}{*}{\textbf{Executed action}}
        & \multicolumn{5}{c}{\textbf{LIBERO}}
        & \multicolumn{2}{c}{\textbf{LIBERO-Plus}} \\
        \cmidrule(lr){3-7}
        \cmidrule(lr){8-9}
        & & Spatial & Object & Goal & Long & Avg.
        & Camera & Robot \\
        \midrule
        \multirow{2}{*}{\textbf{$B \rightarrow R$}}
        & Camera & 96.4 & 99.0 & 97.0 & 93.0 & 96.4
        & 50.2 & \textbf{80.3} \\
        & Wrist  & 86.6 & 99.4 & 90.8 & 56.6 & 83.4
        & \textbf{70.0} & 58.8 \\
        \midrule
        \multirow{2}{*}{\textbf{$R \rightarrow B$}}
        & Camera & 98.4 & 99.4 & 96.2 & 96.8 & 97.7
        & 47.8 & \textbf{84.6} \\
        & Wrist  & 90.6 & 98.6 & 93.2 & 83.8 & 91.6
        & \textbf{73.7} & 69.7 \\
        \bottomrule
    \end{tabular}
\end{table*}

As shown in Table~\ref{tab:libero_plus_full}, Base-only Inference outperforms Full Multi-Expert Inference under Camera perturbations (78.4\% vs.\ 75.4\%), whereas Full Multi-Expert Inference performs better under Robot perturbations (88.4\% vs.\ 81.2\%). 
To further analyze this difference, we execute camera-frame and wrist-frame actions under the $B \rightarrow R$ and $R \rightarrow B$ attention graphs. As shown in Table~\ref{tab:action_frame_attention}, wrist-frame actions consistently outperform camera-frame actions under Camera perturbations for both attention graphs, while the opposite trend is observed under Robot perturbations. Camera perturbations alter the observation--action correspondence associated with the fixed camera, whereas the wrist frame maintains a fixed relationship with the robot end effector, making wrist-frame action predictions relatively stable. Robot perturbations change the robot's initial state and the action trajectory required to complete the task, which also changes the wrist-camera observation. In contrast, the fixed camera continues to provide stable global visual information, making camera-frame action predictions relatively stable. 
Consequently, the two perturbation types affect the action reference frames differently, reversing the relative advantage of camera-frame and wrist-frame actions.

These results further explain the performance difference between the two inference modes. In the $R \rightarrow B$ graph used by Full Multi-Expert Inference, the Base expert attends to action features from the camera and wrist experts and uses them for base-frame action prediction. Under Camera perturbations, the success rate of camera-frame actions decreases substantially, and the useful information provided by the wrist expert is insufficient to fully compensate for the degradation of the camera expert, limiting the overall benefit of multi-expert interaction. Under Robot perturbations, camera-frame actions maintain a high success rate, indicating that the camera expert provides reliable motion information. In contrast, Base-only Inference uses the $B \rightarrow R$ graph, in which the Base expert retains all visual inputs while its action prediction remains independent of view-specific action features. This allows Base-only Inference to avoid the influence of the camera expert under Camera perturbations, but it also does not directly use the additional motion information provided by the camera expert under Robot perturbations, leading to the different behavior of the two inference modes across the two settings. The same reversal between camera-frame and wrist-frame actions appears under both attention graphs, further indicating that this difference arises from how the action reference frames adapt to Camera and Robot perturbations.

\subsection{Complete Results on VLABench and MimicGen}
The complete task-wise results on \textbf{VLABench} are presented in Table~\ref{tab:vlabench_taskwise}. Both inference modes outperform the reproduced $\pi_{0.5}$ baseline overall, with Full mode performing slightly better. On Select Chemistry Tube, Base Only and Full improve the success rate from 6\% to 78\% and 70\%, respectively. Both modes also improve success rates and progress scores on Select Book, Select Mahjong, and Select Poker, while their relative strengths vary somewhat across tasks. For Select Chemistry Tube and Select Toy, the target entities or distance criteria used by IS do not fully align with the task success conditions. Their IS values can therefore serve as supplementary references but should not be used alone to assess the policy's understanding of the target.

For \textbf{MimicGen}, we provide a breakdown of performance across the ten Panda D0 tasks in Table~\ref{tab:mimicgen_results}. Both inference modes achieve higher average success rates than the baseline, with Full mode performing slightly better. On the multi-stage Coffee Preparation task, Base Only and Full improve the success rate from 32\% to 65\% and 72\%, respectively. On Threading, which requires precise alignment and insertion, both modes improve the success rate from 35\% to 42\%. These results demonstrate improvements on both long-horizon and precision-demanding manipulation tasks.

\label{app:another_bench}
\begin{table*}[htpb]
    \centering
    \caption{\textbf{Performance comparison on MimicGen D0 distribution.}
    We report the success rate (\%) for each task and the average performance.$^{*}$ denotes results reproduced by us.}
    \label{tab:mimicgen_results}
    \normalsize
    \setlength{\tabcolsep}{4pt}
    \renewcommand{\arraystretch}{1.08}
    \begin{tabular}{@{}lccccccccccc@{}}
        \toprule
        \textbf{Method}
        & \textbf{Coffee}
        & \makecell{\textbf{Coffee}\\\textbf{Prep.}}
        & \textbf{Hammer}
        & \textbf{Kitchen}
        & \textbf{Mug}
        & \textbf{Square}
        & \textbf{Stack}
        & \makecell{\textbf{Stack}\\\textbf{3}}
        & \textbf{Thread}
        & \makecell{\textbf{3P}\\\textbf{Assembly}}
        & \textbf{Avg.} \\
        \midrule
        $\pi_{0.5}^{*}$
        & 71 & 32 & 97 & 99 & 44 & 58 & 97 & 79 & 35 & 32 & 64.4 \\
        \textbf{Ours} (Base Only)
        & \textbf{85} & 65 & \textbf{100} & \textbf{100} & 48 & 71 & \textbf{100} & 78 & \textbf{42} & \textbf{38} & 72.7 \\
        \textbf{Ours} (Full)
        & 84 & \textbf{72} & \textbf{100} & \textbf{100} & \textbf{56} & \textbf{78} & 98 & \textbf{90} & \textbf{42} & 33 & \textbf{75.3} \\
        \bottomrule
    \end{tabular}
\end{table*}

\begin{table*}[htpb]
    \centering
    \caption{\textbf{Task-wise performance on VLABench.} We report Success Rate (SR), Progress Score (PS), and Intention Score (IS), with all values expressed as percentages. The Average row reports the unweighted mean across the ten tasks. $^{*}$ denotes results reproduced by us.}
    \label{tab:vlabench_taskwise}
    \setlength{\tabcolsep}{4pt}
    \renewcommand{\arraystretch}{1.0}
    \begin{tabular}{@{}lccc@{\hspace{14pt}}ccc@{\hspace{14pt}}ccc@{}}
        \toprule
        \multirow{2}{*}{\textbf{Task}}
        & \multicolumn{3}{c}{$\boldsymbol{\pi}_{0.5}^{*}$}
        & \multicolumn{3}{c}{\textbf{Ours (Base Only)}}
        & \multicolumn{3}{c}{\textbf{Ours (Full)}} \\
        \cmidrule(lr){2-4}
        \cmidrule(lr){5-7}
        \cmidrule(lr){8-10}
        & \textbf{SR} & \textbf{PS} & \textbf{IS}
        & \textbf{SR} & \textbf{PS} & \textbf{IS}
        & \textbf{SR} & \textbf{PS} & \textbf{IS} \\
        \midrule
        Add Condiment & 54 & 75.3 & 98 & 50 & 72.7 & 98 & \textbf{66} & 80.7 & 96 \\
        Insert Flower & 22 & 56 & 86 & 26 & 56 & 84 & \textbf{30} & 63 & 100 \\
        Select Book & 50 & 56 & 76 & 64 & 73 & 88 & \textbf{74} & 82 & 98 \\
        Select Chemistry Tube & 6 & 31 & 2 & \textbf{78} & 87 & 2 & 70 & 85 & 2 \\
        Select Drink & 54 & 66 & 90 & 58 & 71 & 92 & \textbf{68} & 76 & 92 \\
        Select Fruit & 46 & 66 & 94 & \textbf{50} & 71 & 96 & 44 & 71 & 100 \\
        Select Mahjong & 34 & 41 & 92 & 56 & 59 & 92 & \textbf{70} & 76 & 98 \\
        Select Painting & 34 & 34 & 62 & \textbf{44} & 44 & 70 & 42 & 42 & 66 \\
        Select Poker & 74 & 75.3 & 84 & 88 & 90 & 98 & \textbf{92} & 93.3 & 100 \\
        Select Toy & 48 & 70 & 20 & \textbf{54} & 74 & 26 & 50 & 73 & 24 \\
        \midrule
        \textbf{Average} & 42.2 & 57.1 & 70.4 & 56.8 & 69.8 & 74.6 & \textbf{60.6} & \textbf{74.2} & \textbf{77.6} \\
        \bottomrule
    \end{tabular}
\end{table*}


\subsection{Qualitative Results across Simulation Benchmarks}
\label{app:qualitative_results}
\begin{figure*}[htbp]
    \centering
    \includegraphics[width=0.8\textwidth]{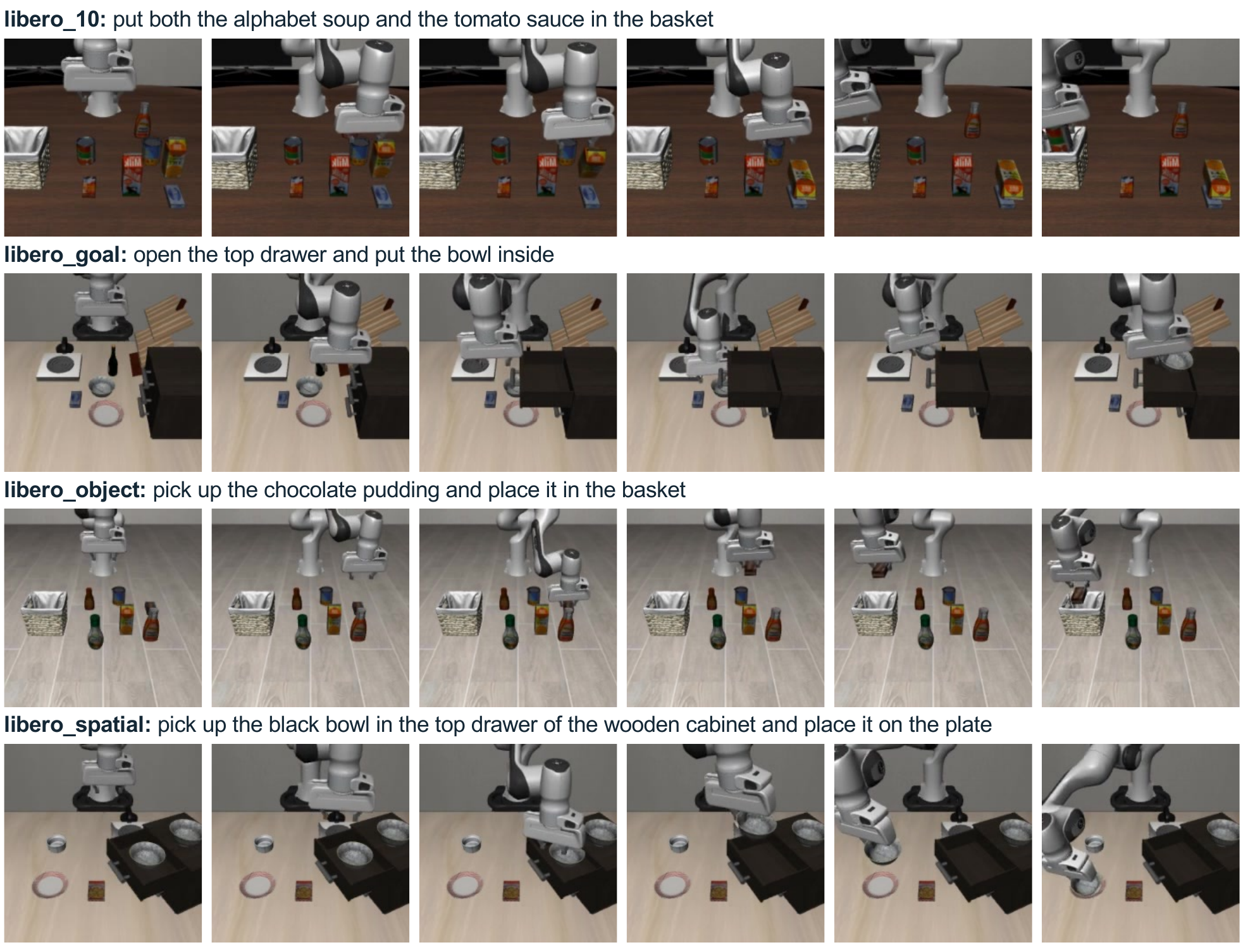}
    \caption{\textbf{Qualitative results on LIBERO.} Successful execution trajectories from the Spatial, Object, Goal, and Long task suites.}
    \label{fig:libero_rollouts}
\end{figure*}

\begin{figure*}[htbp]
    \centering
    \includegraphics[width=\textwidth]{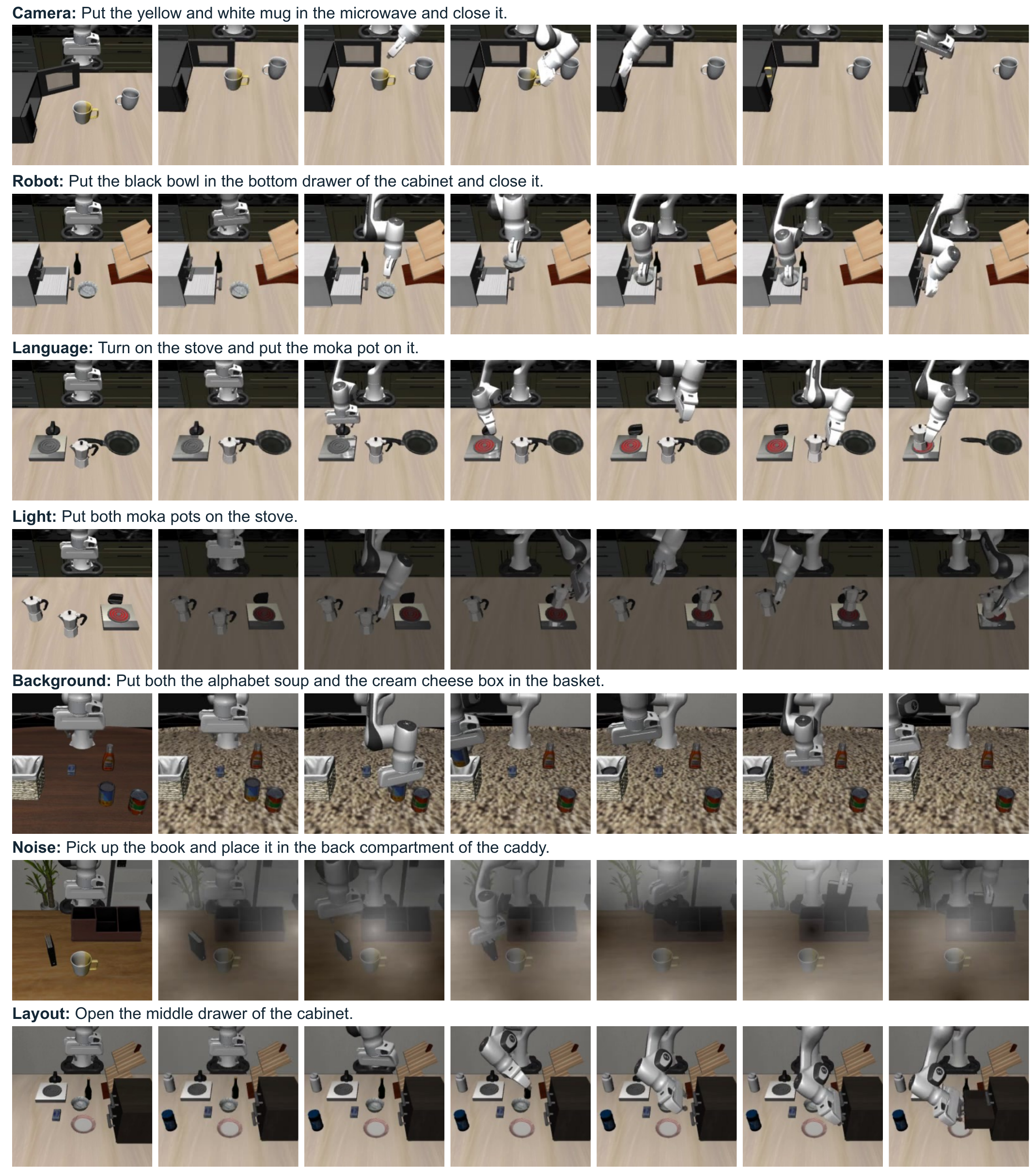}
    \caption{\textbf{Qualitative results on LIBERO-Plus.} Each row shows a successful rollout under one perturbation category: Camera, Robot, Language, Light, Background, Noise, or Layout. The first column presents the initial state from the corresponding unperturbed rollout for comparison, while the remaining six columns show keyframes from the successful perturbed rollout, including its initial frame.}
    \label{fig:libero_plus_rollouts}
\end{figure*}

\begin{figure*}[htbp]
    \centering
    \includegraphics[width=0.65\textwidth]{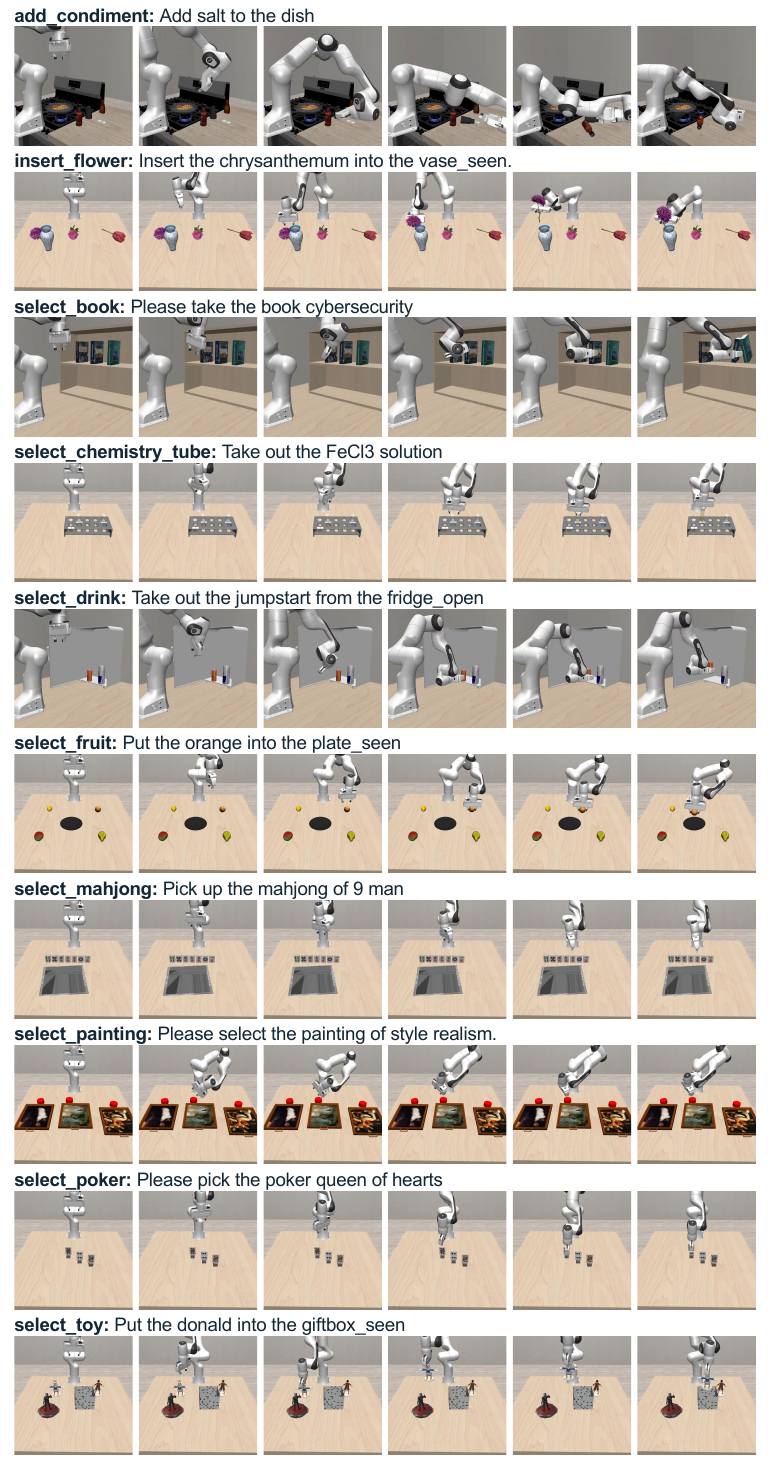}
    \caption{\textbf{Qualitative results on VLABench.} One successful execution trajectory for each of the ten evaluated tasks.}
    \label{fig:vlabench_rollouts}
\end{figure*}

\begin{figure*}[htbp]
    \centering
    \includegraphics[width=\textwidth]{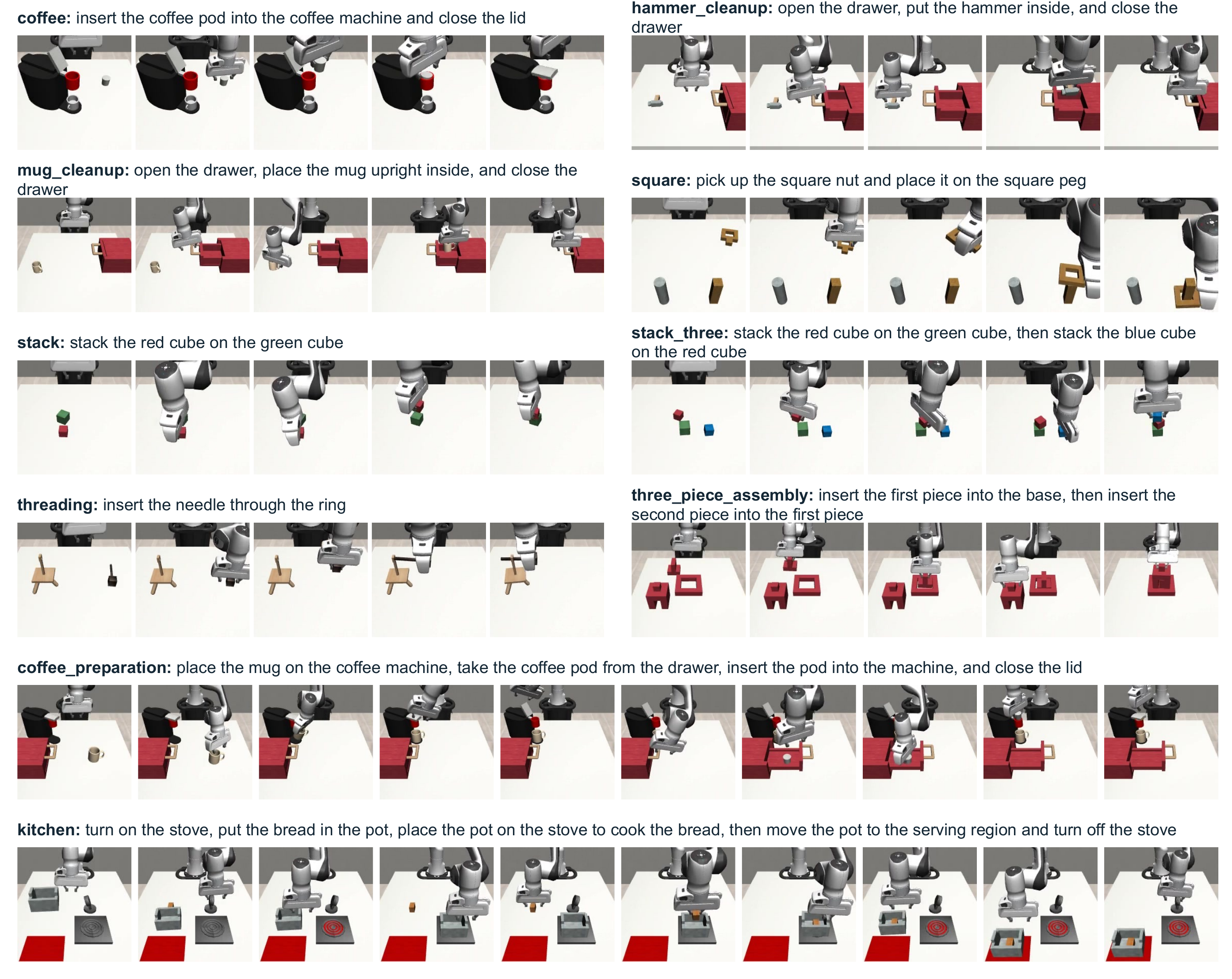}
    \caption{\textbf{Qualitative results on MimicGen.} One successful execution trajectory for each of the ten selected Panda D0 tasks.}
    \label{fig:mimicgen_rollouts}
\end{figure*}

Figures~\ref{fig:libero_rollouts}, \ref{fig:libero_plus_rollouts}, \ref{fig:vlabench_rollouts}, and~\ref{fig:mimicgen_rollouts} present successful execution trajectories of our method across four simulation benchmarks. For LIBERO, we show one trajectory from each of the four task suites; for LIBERO-Plus, one trajectory under each of the seven perturbation types. For VLABench and MimicGen, we show one successful trajectory for every evaluated task. Task names are annotated in the figures, and temporally ordered keyframes illustrate the execution process.

\section{Real-World Experimental Details}
\label{app:real_world}

\subsection{Task Definitions}
\label{app:real_tasks}
We consider three short-horizon tasks and three long-horizon tasks, covering color and spatial understanding, pick-and-place manipulation, precise alignment, tool use, and articulated-object interaction. Table~\ref{tab:real_task_definitions} summarizes the language instructions, task setups, and success criteria. For long-horizon tasks, success requires completion of all specified objectives.

\begin{table*}[t]
\centering
\caption{\textbf{Real-world task definitions.}
We describe the language instruction, task setup, and success criterion
for each short-horizon and long-horizon task.}
\label{tab:real_task_definitions}
\small
\setlength{\tabcolsep}{4pt}
\renewcommand{\arraystretch}{1.15}
\begin{tabularx}{\textwidth}{
    @{}
    >{\raggedright\arraybackslash}p{0.11\textwidth}
    >{\raggedright\arraybackslash}p{0.20\textwidth}
    >{\raggedright\arraybackslash}X
    >{\raggedright\arraybackslash}p{0.25\textwidth}
    @{}
}
\toprule
\textbf{Task} &
\textbf{Language Instruction} &
\textbf{Task Setup and Initialization} &
\textbf{Success Criterion} \\
\midrule
\multicolumn{4}{@{}l}{\textit{Short-horizon tasks}} \\
\midrule

\textbf{Pick Cucumber} &
\textit{``Pick up the cucumber and place it into the blue bowl.''} &
A cucumber and two differently colored bowls are placed on the table.
The robot must identify the blue bowl and place the cucumber inside it.
During data collection, the cucumber's initial position is randomized,
and the positions of the two bowls are randomly exchanged. &
After the gripper releases the cucumber, it lies entirely inside the
blue bowl. \\ \addlinespace[6pt]

\textbf{Stack Blocks} &
\textit{``Place the red block on top of the blue block.''} &
The robot must pick up the red block and stack it on the blue block.
Both blocks' initial positions are randomized without a fixed left--right
ordering, with small variations in their in-plane orientations. &
After release, most of the red block's bottom face rests on the blue
block, and the stack remains stable. Exact centering is not required. \\ \addlinespace[6pt]

\textbf{Hang Mug} &
\textit{``Hang the mug on the mug rack.''} &
The robot must grasp the mug, adjust its orientation, and align its
handle with a hook on the rack. The mug's initial position and handle
orientation are varied during data collection. &
The handle is placed over the hook, and the mug remains stably supported
by the rack after the gripper releases it. \\ 

\midrule
\multicolumn{4}{@{}l}{\textit{Long-horizon tasks}} \\
\midrule

\textbf{Stack Three Cups} &
\textit{``Stack all the cups on the leftmost cup.''} &
Three equally sized cups of different colors are placed separately on
the table. The robot must sequentially stack the other two cups onto the
initially leftmost cup. Cup positions vary within their designated regions,
with variations in relative placement and small orientation perturbations. &
The initially leftmost cup remains at the bottom, and the other two cups
are stacked above it, forming a single stable three-cup stack after release. \\ \addlinespace[6pt]

\textbf{Erase Mark} &
\textit{``Pick up the whiteboard eraser and erase the black X on the whiteboard.''} &
A whiteboard is fixed horizontally on the table, with a black X drawn
on its surface and an eraser placed beside it. The robot must grasp the
eraser, move it to the mark, and perform consecutive wiping motions. &
The robot grasps the eraser and removes most of the black X while the
whiteboard remains stable throughout the operation. \\ \addlinespace[6pt]

\textbf{Fill Drawer} &
\textit{``Place the orange block into the open drawer, and then close the drawer.''} &
The drawer is initially open, and an orange block is placed on the table.
The robot must first place the block inside the drawer and then push
the drawer closed. &
The orange block is inside the drawer, and the drawer is fully closed. \\ 

\bottomrule
\end{tabularx}
\end{table*}

\subsection{Implementation Details}
\label{app:real_implementation}

\begin{figure}[htbp]
    \centering
    \includegraphics[width=0.5\linewidth]{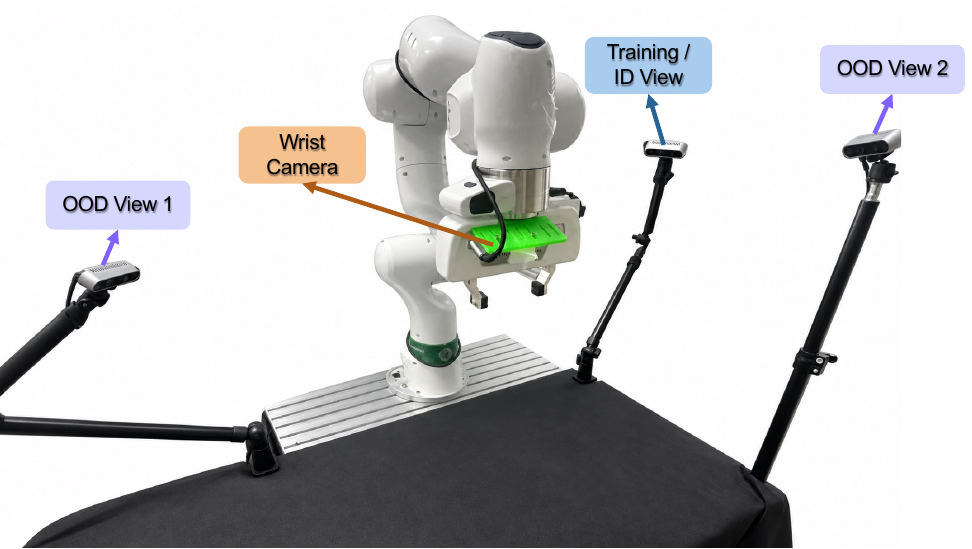}
    \caption{\textbf{Real-world setup.} The training and additional test cameras are indicated.}
    \label{fig:real_world_setup}
\end{figure}

\paragraph{Hardware Setup.}
As shown in Figure~\ref{fig:real_world_setup}, we conduct real-world experiments using a 7-DoF Franka Research 3 (FR3) robot equipped with a parallel-jaw gripper, and collect demonstrations using GELLO teleoperation. Visual observations are captured by Intel RealSense D415 cameras at an RGB resolution of $640\times480$ pixels and 30 FPS. Only RGB streams are used. The setup comprises one wrist-mounted camera and three fixed third-person cameras. The policy always receives two visual inputs: the wrist-mounted view and one selected third-person view. During data collection and in-distribution evaluation, we use the same reference third-person camera together with the wrist-mounted camera. For camera viewpoint shift evaluation, the third-person input is switched to one of the other two fixed cameras, while the wrist-mounted input and all other experimental conditions remain unchanged. Camera roles are annotated in the figure, and detailed viewpoint configurations are provided in Appendix~\ref{app:spatial_shift}.

\paragraph{Camera Calibration.}
To obtain the base-to-camera transformations ${}^{c_i}_{b}T$ used in our method, we calibrate the extrinsics of three fixed cameras and one wrist-mounted camera. We use a ChArUco board and collect images together with the corresponding robot end-effector poses at multiple positions and orientations. Given the camera intrinsics and detected board corners, we estimate the board pose in each camera frame using PnP, then solve for the camera extrinsics using OpenCV hand-eye calibration. Below, $m$ denotes the calibration board frame, and $k$ indexes calibration samples.

For each fixed camera, the board is rigidly attached to the end effector, yielding
\begin{equation}
    {}^{c_i}_{m}T(k)
    =
    {}^{c_i}_{b}T\,
    {}^{b}_{e}T(k)\,
    {}^{e}_{m}T.
\end{equation}
Here, ${}^{c_i}_{m}T(k)$ is estimated from the image, ${}^{b}_{e}T(k)$ is provided by the robot, and ${}^{e}_{m}T$ remains constant throughout calibration. We perform eye-to-hand calibration using these observations to obtain ${}^{c_i}_{b}T$ for each fixed camera.

For the wrist-mounted camera, the board is fixed in the workspace, and the poses satisfy:
\begin{equation}
    {}^{c_w}_{m}T(k)
    =
    {}^{c_w}_{e}T\,
    \bigl({}^{b}_{e}T(k)\bigr)^{-1}
    {}^{b}_{m}T,
\end{equation}
where ${}^{b}_{m}T$ remains constant. Eye-in-hand calibration yields the fixed end-effector-to-camera transformation ${}^{c_w}_{e}T$. At each observation timestep $t$, we compute the base-to-wrist-camera transformation from the current end-effector pose:
\begin{equation}
    {}^{c_w}_{b}T(t)
    =
    {}^{c_w}_{e}T\,
    \bigl({}^{b}_{e}T(t)\bigr)^{-1}.
\end{equation}
These extrinsics are used for the state and action transformations defined in the main text. In our real-world implementation, all actions within a predicted chunk use the wrist-camera frame at the initial observation timestep. The transformation is updated when a new observation is acquired for replanning.

\paragraph{Training and Execution.}
Our method and the reproduced $\pi_{0.5}$ baseline are trained on the same pooled multi-task demonstration dataset using the corresponding language task instructions, and share the same main training hyperparameters. Both models are trained for 30K steps with a global batch size of 64 using AdamW. The learning rate is warmed up to $2.5\times10^{-5}$ over the first 1,000 steps, followed by cosine decay toward $2.5\times10^{-6}$. Demonstration trajectories are sampled at 30 Hz for training.
Both policies predict 7-D Cartesian actions consisting of a 3-D translational offset, a 3-D world-relative rotation vector, and a continuous target gripper-closedness value in $[0,1]$. At each inference step, the policy predicts a 10-step action chunk, with all actions decoded relative to the end-effector pose observed at the beginning of the chunk. The first five actions are executed before acquiring new observations and replanning. The continuous gripper prediction is converted into open/close commands using hysteresis: values at or below $0.3$ trigger opening, values at or above $0.7$ trigger closing, and intermediate values retain the previous gripper state.

\subsection{Spatial Distribution Shift Details}
\label{app:spatial_shift}
\begin{figure*}[htbp]
    \centering
    \includegraphics[width=0.85\textwidth]
    {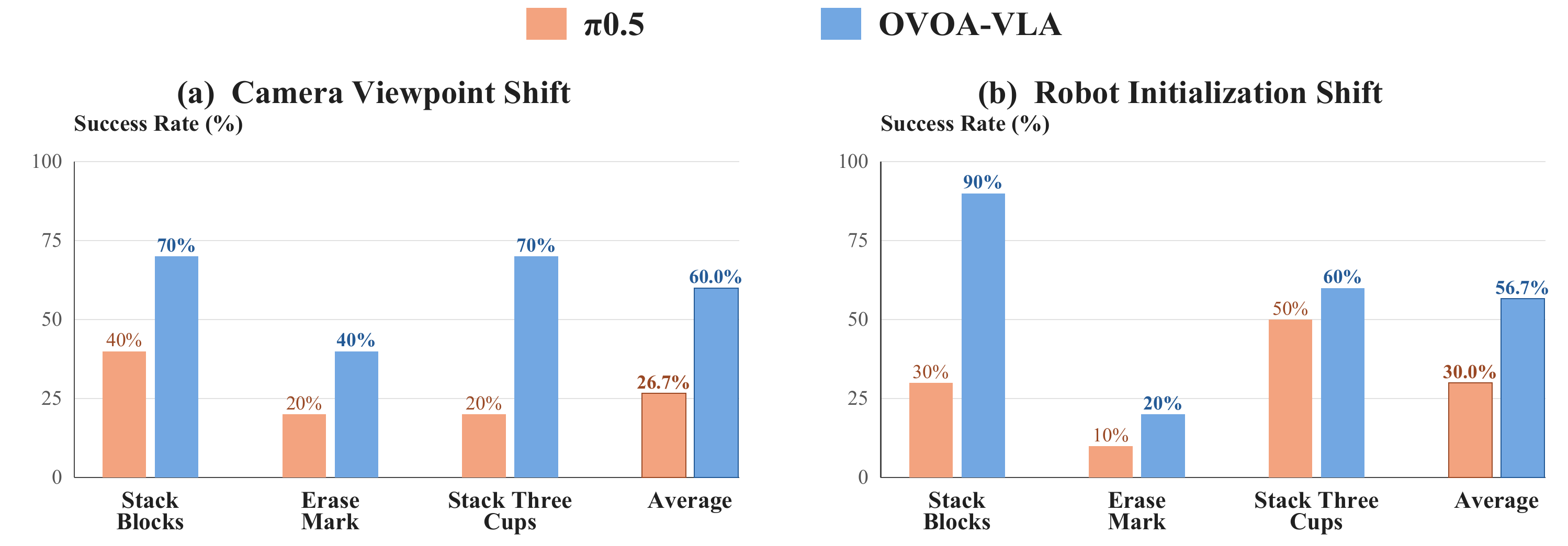}
    \caption{
        \textbf{Detailed zero-shot robustness under spatial distribution shifts.}
        Success rates under camera viewpoint and robot initialization shifts.
        Each task aggregates two OOD configurations with five rollouts per configuration; Average denotes the mean across the three tasks.
    }
    \label{fig:real_world_ood_shift_breakdown}
\end{figure*}

As shown in Figure~\ref{fig:real_world_setup}, training and in-distribution evaluation use the same fixed third-person camera together with the wrist-mounted camera. For the camera viewpoint shift, the training third-person view is replaced with either OOD View 1 or OOD View 2, neither of which is used during demonstration collection or post-training, while the wrist-mounted view remains unchanged. For the robot initialization shift, the robot is first reset to its nominal task-specific configuration, after which an EEF pose offset is applied in the robot base frame. The left configuration applies a translation of $\Delta\mathbf{p}=[0,0.10,0]\,\mathrm{m}$ and a yaw rotation of $+45^\circ$, whereas the right configuration applies $\Delta\mathbf{p}=[0,-0.10,0]\,\mathrm{m}$ and a yaw rotation of $-45^\circ$. All other evaluation conditions remain unchanged. We evaluate Stack Blocks, Erase Mark, and Stack Three Cups under both types of spatial shift, conducting five rollouts for each task and configuration.

As shown in Figure~\ref{fig:real_world_ood_shift_breakdown}, OVOA-VLA improves the average success rate from 26.7\% to 60.0\% under camera viewpoint shifts and from 30.0\% to 56.7\% under robot initialization shifts. It outperforms $\pi_{0.5}$ across all six task--shift combinations, with the largest gains observed on Stack Three Cups under camera viewpoint shifts and Stack Blocks under robot initialization shifts. These results show that the improvement is not limited to a single type of spatial shift.

\subsection{Additional Real-World Qualitative Results}
\begin{figure*}[htbp]
    \centering
    \includegraphics[width=\textwidth]{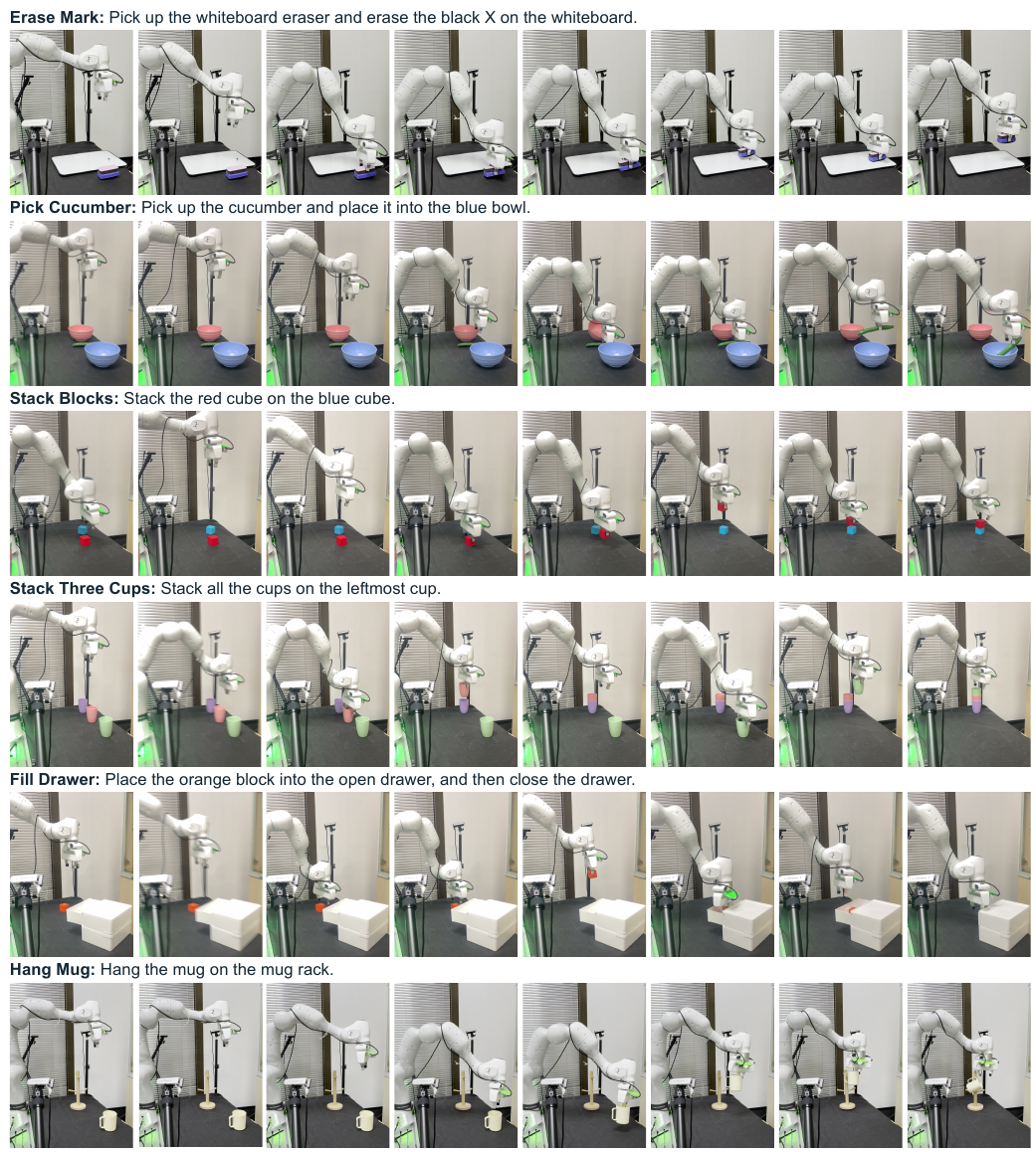}
    \caption{\textbf{Qualitative results under in-distribution evaluation.}
    Representative successful OVOA-VLA rollouts across the six evaluated real-world tasks.}
    \label{fig:real_world_id_rollouts}
\end{figure*}

\begin{figure*}[htbp]
    \centering
    \includegraphics[width=0.95\textwidth]{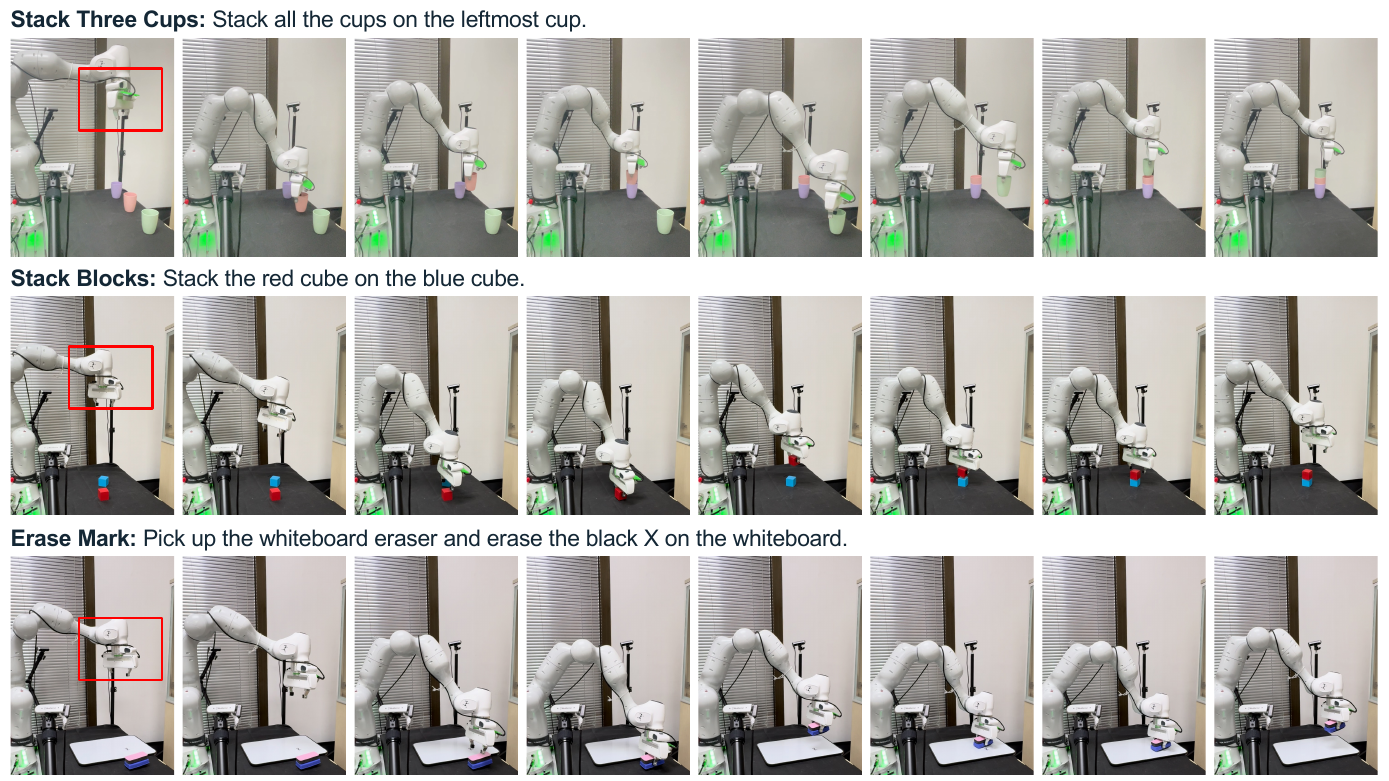}
    \caption{\textbf{Qualitative results under robot initialization shifts.}
        Representative successful OVOA-VLA rollouts on Stack Blocks, Erase Mark, and Stack Three Cups. The red box in the first frame of each rollout highlights the perturbed initial EEF pose.}
    \label{fig:real_world_robot_ood_rollouts}
\end{figure*}

We provide additional qualitative examples to complement the quantitative real-world evaluation. Figure~\ref{fig:real_world_id_rollouts} presents successful executions across all six tasks under the in-distribution setting, while Figure~\ref{fig:real_world_robot_ood_rollouts} shows successful executions from perturbed robot initializations.


